\documentclass{article}
\usepackage{iclr2027_conference,times}

\usepackage{amsmath,amsfonts,bm}

\def\eqref#1{equation~\ref{#1}}

\def\1{\bm{1}}

\DeclareMathAlphabet{\mathsfit}{\encodingdefault}{\sfdefault}{m}{sl}
\SetMathAlphabet{\mathsfit}{bold}{\encodingdefault}{\sfdefault}{bx}{n}

\usepackage{booktabs}
\usepackage{graphicx}
\usepackage{colortbl}
\usepackage{xcolor}
\usepackage{soul}
\usepackage{hyperref}
\usepackage{url}
\usepackage{placeins}
\usepackage{algorithm}
\usepackage{algpseudocode}
\usepackage{float}
\usepackage[most]{tcolorbox}

\definecolor{gainGreen}{RGB}{24,145,91}
\definecolor{lossRed}{RGB}{205,70,80}
\definecolor{bestLavender}{RGB}{224,218,244}
\definecolor{secondSand}{RGB}{246,235,210}
\definecolor{groupGray}{RGB}{244,244,244}
\definecolor{authorHighlight}{RGB}{255,244,153}
\definecolor{paperLinkBlue}{RGB}{30,78,135}
\hypersetup{
  colorlinks=true,
  linkcolor=paperLinkBlue,
  citecolor=paperLinkBlue,
  urlcolor=paperLinkBlue
}
\newtcblisting{promptbox}{
  enhanced,
  breakable,
  listing only,
  colback=groupGray,
  colframe=black!35,
  boxrule=0.45pt,
  arc=1mm,
  left=1.5mm,
  right=1.5mm,
  top=1mm,
  bottom=1mm,
  before skip=5pt,
  after skip=7pt,
  listing options={
    basicstyle=\ttfamily\footnotesize,
    breaklines=true,
    breakatwhitespace=true,
    breakautoindent=false,
    breakindent=0pt,
    columns=fullflexible,
    keepspaces=true,
    tabsize=2,
    showstringspaces=false
  }
}
\newcommand{\promptheading}[1]{%
  \par\medskip\noindent\textbf{#1}\par\nobreak\vspace{3pt}%
}
\newcommand{\gain}[1]{\textcolor{gainGreen}{\scriptsize$\uparrow$\,#1}}
\newcommand{\loss}[1]{\textcolor{lossRed}{\scriptsize$\downarrow$\,#1}}
\newcommand{\roundgain}[1]{\textcolor{gainGreen}{\scriptsize$\downarrow$\,#1}}
\newcommand{\roundloss}[1]{\textcolor{lossRed}{\scriptsize$\uparrow$\,#1}}

\title{Dense Is Not Enough: Hierarchical Supervision Allocation for Long-Horizon On-Policy Distillation}

\author{\textbf{Yuhao Sun\textsuperscript{1,3,*}}
\quad
\textbf{Binrui Wu\textsuperscript{2,*}}
\quad
\textbf{Zhuoer Xu\textsuperscript{1,\ensuremath{\dagger}}}
\quad
\textbf{Ming Wen\textsuperscript{1}}
\\
\textbf{Haoxiang Xu\textsuperscript{3}}
\quad
\textbf{Bin Chen\textsuperscript{4}}
\quad
\textbf{Yan Lin\textsuperscript{5}}
\quad
\textbf{Qianzijing Zhang\textsuperscript{3}}
\\[2mm]
\textsuperscript{1}Ant Group
\quad
\textsuperscript{2}Alibaba International Digital Commerce Group
\\
\textsuperscript{3}University of Science and Technology of China
\quad
\textsuperscript{4}Peking University
\\
\textsuperscript{5}University of Electronic Science and Technology of China
\\[1mm]
\textsuperscript{*}Equal contribution
\qquad
\textsuperscript{\ensuremath{\dagger}}Corresponding author
}

\iclrfinalcopy

\begin{document}

\maketitle

\begin{abstract}
On-policy distillation (OPD) transfers the capabilities of a large language model to a smaller student by providing teacher supervision on the student's own rollouts.  In long-horizon agentic tasks, however, uniform token-level matching can allocate supervision poorly: a large local discrepancy need not improve future behavior, while consequential guidance may be beyond the current student's reach or fail to persist without privileged input.  We formulate long-horizon OPD as \emph{hierarchical supervision allocation} and argue that productive guidance lies at the intersection of \emph{future utility} and \emph{current learnability}.  Crucially, this intersection evolves as the student learns.  Based on this principle, we propose \textbf{LENS-OPD}, a coarse-to-fine framework that organizes supervision through \textsc{Locate}, \textsc{Validate}, and \textsc{Refine}.  \textsc{Locate} adapts trajectory exposure to the student's evolving competence and proposes a candidate decision for intervention.  \textsc{Validate} tests whether teacher guidance at that decision improves the same student's subsequent behavior.  \textsc{Refine} internalizes the beneficial guided behavior into the deployable policy and concentrates token-level supervision on decisive teacher--student conflicts within the validated turn.  These stages are nested: each finer allocation is conditioned on the coarser decision, rather than being optimized as an independent importance score.  Experiments across multiple long-horizon agent benchmarks and student--teacher configurations show that LENS-OPD consistently improves task performance over vanilla OPD and strong curriculum- and selection-based baselines.  Our results suggest that effective long-horizon distillation requires teaching at the right depth, the right decision, and the right token.
\end{abstract}

\section{Introduction}
\label{sec:introduction}

On-policy distillation (OPD) transfers the capabilities of a large language model to a smaller
student~\citep{gu2023minillm,agarwal2024policy}.  Unlike offline distillation on fixed teacher
demonstrations, OPD collects student rollouts and queries a frozen teacher on the contexts the student
actually visits.  It therefore combines on-policy data with dense distributional supervision and
reduces exposure bias~\citep{DBLP:conf/aistats/RossGB11,DBLP:conf/nips/BengioVJS15}.  This recipe is
well suited to single-turn generation, but is incomplete for long-horizon agentic tasks, where each
action can change later observations and decisions~\citep{DBLP:conf/iclr/ShridharYCBTH21,
DBLP:conf/nips/Yao0YN22,DBLP:conf/emnlp/WangJCA22}.  Effective supervision must therefore decide how
far the student should explore, which decision to correct, and what part of the correction it can
absorb.

Recent methods approach this problem from two directions.  \emph{Trajectory-control methods}
regulate rollout depth or alternate teacher and student control to reduce distribution drift and
compounding errors~\citep{wang2026tcod,li2026policy,zhou2026turnopd}.
\emph{Selective-supervision methods} allocate teacher guidance using disagreement, confidence,
environment feedback, or future continuations~\citep{zhou2026sage,chen2026look}.  Both improve upon
uniform OPD, but neither jointly determines whether a teacher signal has downstream utility and is
learnable by the current student.  Local disagreement cannot distinguish these cases, and a
global curriculum cannot determine which intervention is productive at a particular state.

\begin{figure}[t]
    \centering
    \includegraphics[width=0.97\linewidth]{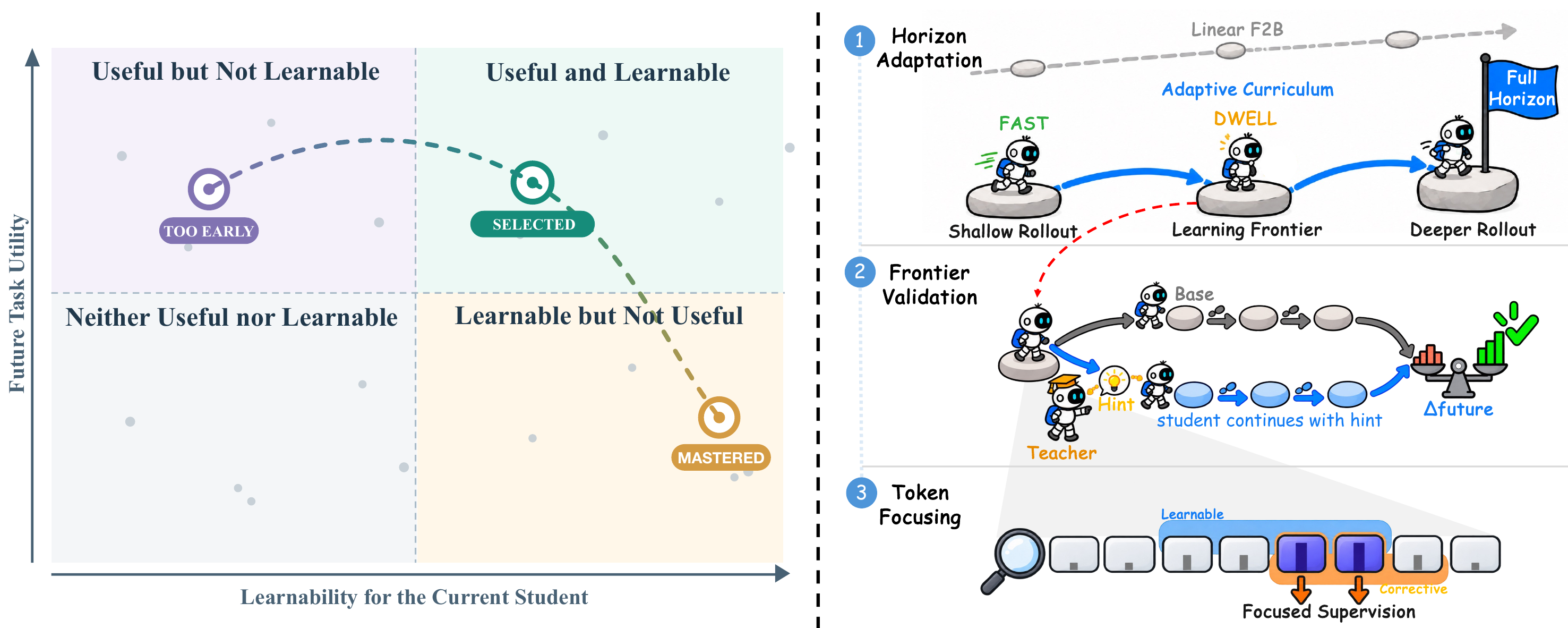}
    \caption{Overview of LENS-OPD. Productive guidance combines future utility and current learnability, while Locate--Validate--Refine allocates supervision from trajectories to turns and tokens.}
    \label{fig:lens-overview}
    \vspace{-4mm}
\end{figure}

We therefore view long-horizon OPD as \emph{hierarchical supervision allocation}.  Our key insight
is that productive guidance lies at the intersection of \emph{future utility} and \emph{current
learnability}, as illustrated in Figure~\ref{fig:lens-overview}.  Future utility asks whether an
intervention improves the student's subsequent behavior; current learnability asks whether the
student can realize that improvement under guidance and internalize it into its policy.  Neither
condition can be decided by token-level discrepancy alone.  Moreover, the allocation decisions are
nested: trajectory exposure determines which states can be visited, turn-level validation determines
which intervention is worth learning from, and token-level refinement determines where the accepted
signal should exert its learning pressure.  This productive region also evolves throughout training.
Guidance that is initially beyond the student's reach may later become learnable, while previously
useful supervision may become redundant once absorbed.  Effective OPD must therefore continually
adapt both what the student explores and where it concentrates its learning capacity.

Based on this view, we propose \textbf{LENS-OPD}, a coarse-to-fine framework that allocates
supervision through \textsc{Locate}, \textsc{Validate}, and \textsc{Refine}.  \textsc{Locate} first
adapts the rollout horizon to the student's evolving competence, then uses local discrepancy only to
propose a candidate decision rather than treating it as evidence of usefulness.  \textsc{Validate}
restores the same state and compares paired continuations generated by the same student with and
without a short teacher hint.  Because the student retains control in both branches, this comparison
tests whether the guidance can improve its own future behavior instead of whether the teacher can
execute a better action.  For guidance that passes this test, \textsc{Refine} transfers the induced
behavior into the hint-free policy and reallocates token-level supervision toward decisive
teacher--student conflicts.  Each finer allocation is thus conditioned on the preceding coarser
decision, progressively narrowing supervision from the right trajectory depth to the right decision
and finally the right tokens.

In summary, this work makes three contributions.  First, we formulate long-horizon OPD as
hierarchical supervision allocation and identify future utility and current learnability as the two
conditions for productive guidance.  Second, we introduce LENS-OPD, which coordinates trajectory,
turn, and token supervision through Locate--Validate--Refine.  Third, experiments across long-horizon
agent benchmarks, student scales, and model families show consistent gains, supported by analyses
of adaptive supervision dynamics and component ablations.

\section{Related Work}
\label{sec:related-work}

\subsection{Knowledge Distillation and On-Policy Distillation}

Knowledge distillation~\citep{hinton2015distilling} transfers teacher behavior
through soft output distributions. Sequence-level distillation~\citep{kim2016sequence}
trains students on teacher-generated sequences. For pretrained Transformers, distillation
also uses hidden-state alignment~\citep{sanh2019distilbert,sun2019patient,jiao2020tinybert,sun2020mobilebert}
and attention-relation matching~\citep{wang2020minilm,wang2021minilmv2}.
On-policy distillation~\citep{gu2023minillm,agarwal2024policy}
supervises students on their own generated prefixes, reducing the mismatch between teacher-generated
training contexts and student-generated inference contexts. On-policy self-distillation~\citep{zhao2026self,shenfeld2026self}
further derives teacher supervision from the same model conditioned on privileged information,
transferring this guidance to a student that operates without such information.

\subsection{Adaptive Supervision in On-Policy Distillation}

Adaptive supervision in single-response generation adjusts the selection, strength, and form
of token-level feedback. Teacher verification and entropy-based sampling~\citep{huang2026selectkd,ko2026scaling}
determine where supervision is applied, while uncertainty--disagreement analysis and
teachability-aware selection~\citep{xu2026tip,wang2026not} distinguish informative corrections
from locally incompatible teacher signals. Decision--evidence relations and reasoning-progress
estimates~\citep{xiao2026finding,yang2026beyond} also guide supervision allocation.
Weighting strategies~\citep{li2026filter,xie2026position} account for response reliability
and accumulated prefix mismatch. Position-dependent objectives~\citep{jin2026entropy,jia2026asymmetric}
adapt supervision to teacher uncertainty or the sign of token-level advantage.
Multi-turn supervision additionally controls state visitation and allocates feedback across decisions. Trajectory curricula
and teacher--student control strategies~\citep{wang2026tcod,li2026policy,xia2026dash} regulate
student exposure by progressively extending rollout horizons, annealing teacher intervention,
or switching executors in response to accumulated discrepancy and recovery signals. Turn-aware
budgeting and supervision schedules~\citep{zhou2026turnopd,tan2026atod} further adapt rollout
depth, redistribute learning effort across turns, and balance distillation with reward-driven
optimization. Selective intervention and confidence-based weighting~\citep{zhou2026sage,tan2026atod}
use environment feedback, teacher judgment, or model uncertainty and disagreement to determine
where guidance is needed and how strongly it should contribute. Future trajectory
validation~\citep{chen2026look} tests whether teacher guidance yields more favorable downstream
distillation signals in subsequent student continuations. Building on these directions, LENS-OPD conditions
both guidance internalization and within-turn token refinement on a paired test of subsequent
student behavior, within a trajectory curriculum that adapts to the student's evolving competence.


\section{Effective Supervision as a Dynamic Frontier}
\label{sec:problem}

\subsection{Agentic On-Policy Distillation}

We consider a student policy $\pi_\theta$ interacting with an environment for at most $T$ turns.
At turn $t$, it observes $o_t$, conditions on
$h_t=(x,o_1,y_1,\ldots,o_t)$, samples the response $y_t\sim\pi_\theta(\cdot\mid h_t)$,
and executes its action to obtain $o_{t+1}$.  A frozen rollout snapshot
$\bar\pi_\theta$ induces the training trajectory
$\tau=(h_1,y_1,\ldots,h_T,y_T)$, and a frozen teacher $\pi_{\mathrm T}$ is evaluated
on the same student-visited token contexts.  Let $c_{t,i}=(h_t,y_{t,<i})$ denote the context for
predicting token $y_{t,i}$, and let $m_{t,i}\in\{0,1\}$ indicate whether that token is included in
the training loss.  Standard OPD minimizes
\begin{equation}
    \mathcal L_{\mathrm{OPD}}(\theta)
    =
    \mathbb E_{\tau\sim\bar\pi_\theta}\!\left[
    \frac{\sum_{t,i}m_{t,i}
    D_{\mathrm{KL}}\!\left(\pi_\theta(\cdot\mid c_{t,i})
    \,\|\,\pi_{\mathrm T}(\cdot\mid c_{t,i})\right)}
    {\sum_{t,i}m_{t,i}}\right].
    \label{eq:agent-opd}
\end{equation}
Although Equation~\ref{eq:agent-opd} is on-policy with respect to the visited contexts, it weights
all masked tokens uniformly, even though supervision differs in both future utility and current
learnability.  We call the evolving set of signals satisfying both conditions the \emph{effective
supervision frontier}.

\subsection{Empirical Diagnostics}

\begin{figure}[t]
  \centering
  \makebox[\textwidth][c]{%
  \begin{minipage}[t]{0.285\textwidth}
    \centering
    \begin{tabular}{@{}c@{}}
      \includegraphics[height=0.155\textheight]{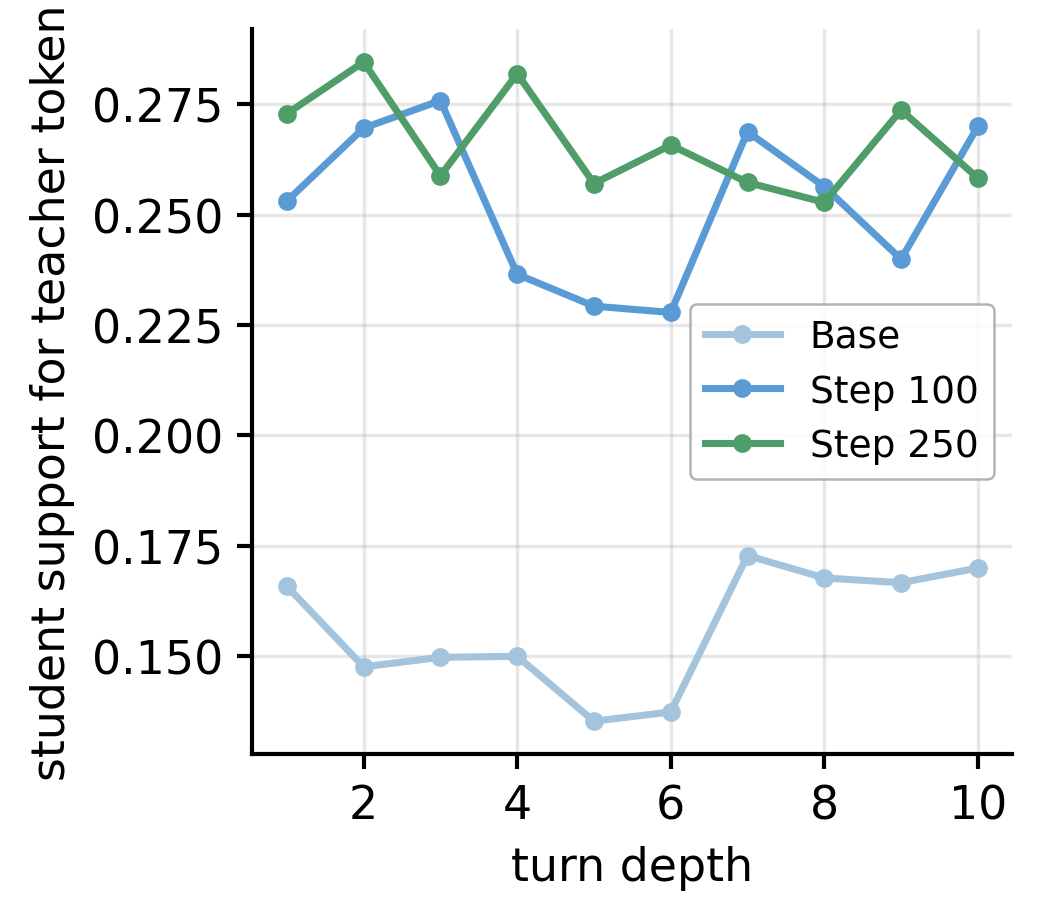} \\[-2mm]
      {\footnotesize (a) Support evolves with training}
    \end{tabular}
  \end{minipage}\hspace{0.015\textwidth}%
  \begin{minipage}[t]{0.285\textwidth}
    \centering
    \begin{tabular}{@{}c@{}}
      \includegraphics[height=0.155\textheight]{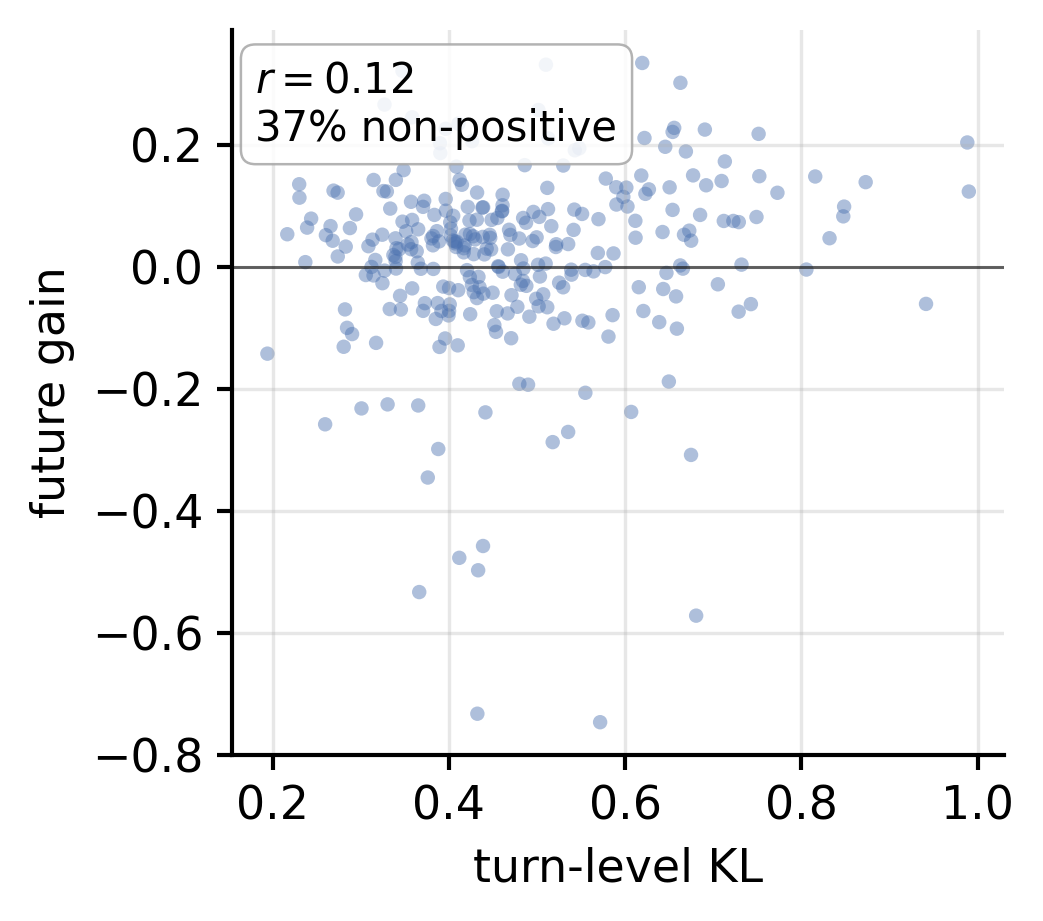} \\[-2mm]
      {\footnotesize (b) KL versus future gain}
    \end{tabular}
  \end{minipage}\hspace{0.015\textwidth}%
  \begin{minipage}[t]{0.285\textwidth}
    \centering
    \begin{tabular}{@{}c@{}}
      \includegraphics[height=0.155\textheight]{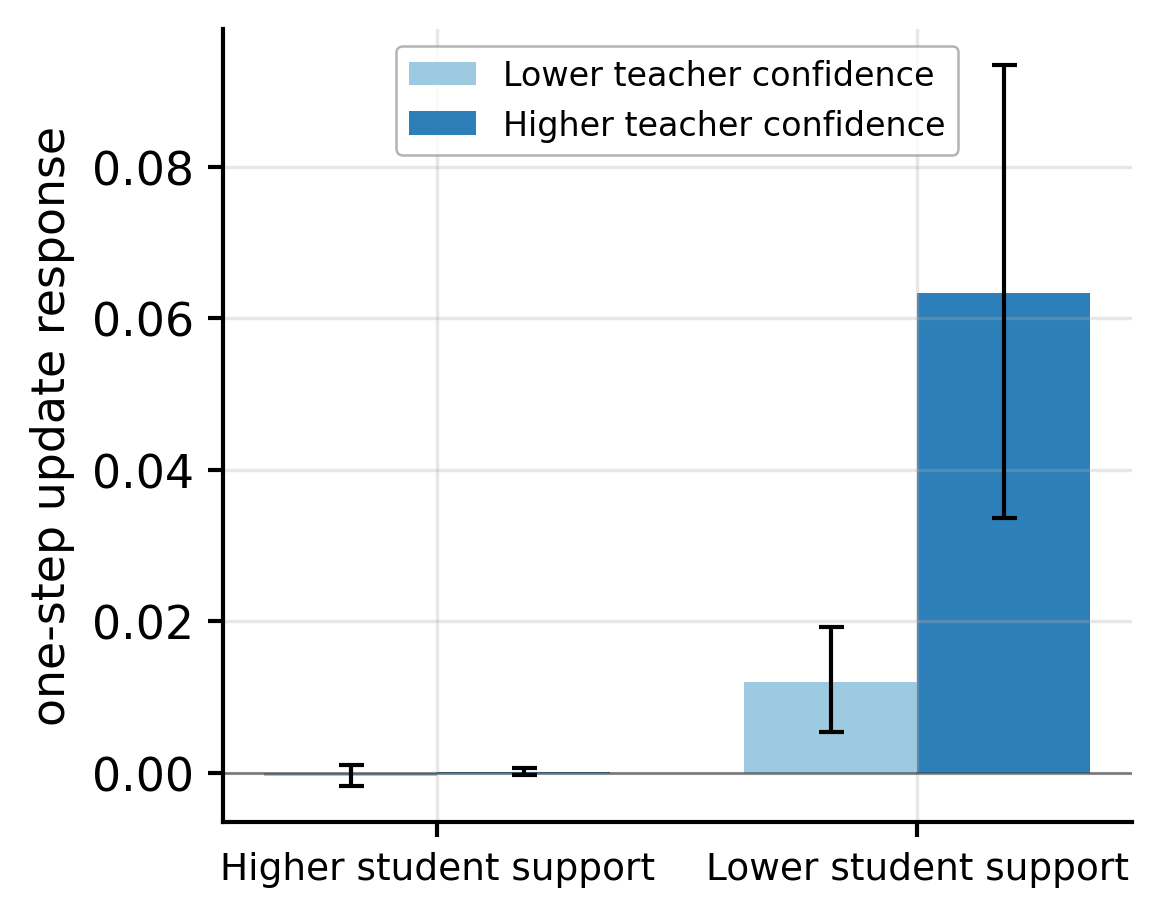} \\[-2mm]
      {\footnotesize (c) Token update response}
    \end{tabular}
  \end{minipage}%
  }
  \caption{\textbf{Diagnostics of hierarchical supervision allocation.}
  (a) Student support for teacher-preferred tokens at disagreement positions changes markedly with
  training.  (b) Turn-level KL is a weak proxy for the future benefit of guidance, with 37\% of
  valid forks producing non-positive gain.  (c) Within future-beneficial turns, one-step corrective
  response is largest when the student assigns lower support and the teacher expresses higher
  confidence.  Error bars in (c) are 95\% bootstrap confidence intervals clustered by fork.}
  \label{fig:motivation}
\end{figure}

We conduct three controlled diagnostics on frozen student checkpoints, summarized in
Figure~\ref{fig:motivation}.  These probes examine how student support changes during training,
whether local discrepancy predicts turn-level utility, and where update signal concentrates within
future-beneficial turns.

\paragraph{Student support evolves with training.}
We first examine positions where the student and teacher prefer different tokens.  At each such
position, student support is the clipped probability ratio between the teacher-preferred token and
the student's own top token.  Figure~\ref{fig:motivation}(a) compares this quantity across turn
depths for the initial student and two later checkpoints.  Support rises substantially after
training and continues to shift between checkpoints, while its profile remains broadly consistent
across depths.  Thus, whether a teacher correction lies within the student's current support cannot
be fixed in advance; supervision exposure must track the evolving student.

\paragraph{Local discrepancy does not determine future utility.}
We next sample candidate turns across the local-KL range and, from each restored state, compare
three-turn continuations produced by the same frozen student with and without a one-sentence teacher
cue.  We define future gain $U_t\equiv\Delta_t^{\mathrm{future}}$ as the difference in teacher
log-likelihood between the two continuations.  Across 286 valid forks, local KL is only weakly
associated with $U_t$, and 37\% of
interventions yield non-positive gain (Figure~\ref{fig:motivation}(b)).  Moreover, the maximum-KL
turn is also the highest-gain turn in only one third of the trajectories.  Local discrepancy can
therefore propose a candidate decision, but cannot determine whether its guidance will improve the
future.

\paragraph{Future utility does not imply uniform token actionability.}
Finally, among forks with positive future gain, we partition disagreement tokens by student support
and teacher confidence.  Starting from the same checkpoint, we apply one fixed OPD update and
measure the increase in the student's log-probability of the teacher-preferred token.  As shown in
Figure~\ref{fig:motivation}(c), the response concentrates where student support is lower and teacher
confidence is higher, reaching roughly five times the largest response in the other groups.  Thus,
future validation determines whether a turn provides productive guidance, but does not make all of
its tokens equally actionable.

Together, these diagnostics motivate hierarchical rather than global supervision selection:
trajectory exposure should adapt online, local KL should only nominate a candidate turn, and paired
future validation should determine whether its guidance is productive.  Only validated guidance is
then refined at the behavior and token levels.

\section{LENS-OPD}
\label{sec:method}

\subsection{Framework Overview}

Let $\pi_\theta$ be the trainable student, $\bar\pi_\theta$ the frozen snapshot used for
rollout and target construction, and $\pi_{\mathrm T}$ a frozen teacher.  LENS-OPD allocates
supervision at three nested resolutions.  At the trajectory level, a gap-adaptive curriculum controls
how deeply $\bar\pi_\theta$ interacts with the environment.  Within the resulting rollout, a cheap
turn-level discrepancy proposes one decision for closer inspection.  A paired future test then
determines whether guidance at the proposed turn both improves the future and can be realized by the
current student.  Only an accepted turn activates two complementary mechanisms for absorption:
guidance internalization transfers the hint-induced behavior into a hint-free student, while
confident-disagreement token focusing reallocates ordinary OPD supervision toward decisive
corrective conflicts in the original on-policy response.

\subsection{Gap-Adaptive Horizon Curriculum}

LENS-OPD retains a stable Forward-to-Backward curriculum clock~\citep{wang2026tcod}, but adapts its pace
to competence gained at the current depth.  The controller maintains the rollout horizon $K$, a
continuous clock $e$, a nominal residence budget $\eta$, and a maximum horizon $K_{\max}$.

\paragraph{Frontier competence.}
Consider an episode that reaches the current cap $K$.  At content-token position $i$ of its frontier
turn, let $v_i^{\mathrm T}$ and $v_i^{\mathrm S}$ denote the teacher and student top-1 tokens under
the same student-generated context.  We define the soft support of the teacher token as
\begin{equation}
    a_i
    =\min\!\left(
      1,
      \frac{\bar\pi_\theta(v_i^{\mathrm T}\mid c_{K,i})}
           {\bar\pi_\theta(v_i^{\mathrm S}\mid c_{K,i})}
      \right),
    \label{eq:frontier-soft-support}
\end{equation}
and average it over valid content positions to obtain episode competence $c_{b,K}$.  Across episodes
that reach $K$, the batch-level frontier competence is
$C_{n,K}=\operatorname{Median}_{b\in\mathcal B_n^{\mathrm{valid}}}c_{b,K}$.
\paragraph{Gap-adaptive pace.}
Upon entering depth $K$, the first valid batch records the entry gap
$G_K^{\mathrm{entry}}=1-C_{\mathrm{entry},K}$.  For a later batch, let
$G_{n,K}=1-C_{n,K}$.  LENS advances the curriculum clock at rate
\begin{equation}
    v_{n,K}
    =\operatorname{clip}\!\left(
      \frac{G_K^{\mathrm{entry}}+\epsilon}
           {G_{n,K}+\epsilon},
      0.5,1.5
      \right),
    \qquad
    e\leftarrow e+v_{n,K}.
    \label{eq:gap-adaptive-pace}
\end{equation}
When $e\geq\eta$, the controller sets
$K\leftarrow\min(K+1,K_{\max})$ and carries the fractional remainder
$e\leftarrow e-\eta$ into the next depth.  The clock therefore accelerates as the student closes the
gap at the current frontier and slows when the gap expands.

\subsection{Turn Proposal and Future Validation}

The horizon controller determines which states are exposed, but it does not designate an
intervention turn.  For each valid turn $t\leq K$, we compute the mean content-token divergence
\begin{equation}
    D_t
    =\frac{1}{|\mathcal C_t|}
      \sum_{i\in\mathcal C_t}
      D_{\mathrm{KL}}\!\left(
        \bar\pi_\theta(\cdot\mid c_{t,i})
        \,\|\,
        \pi_{\mathrm T}(\cdot\mid c_{t,i})
      \right),
    \qquad
    t^\star=\arg\max_{t\leq K}D_t.
    \label{eq:top-kl-proposal}
\end{equation}
We use $D_t$ only as a low-cost, high-recall proposal signal.  As shown in
Figure~\ref{fig:motivation}(b), local divergence is weakly associated with downstream benefit;
thus, $t^\star$ identifies where to inspect, not where to supervise.  We need test whether
intervening at $t^\star$ improves the same student's subsequent behavior.

\paragraph{Paired future validation.}
Rather than replacing the student's proposed action or handing execution to the teacher as in
teacher-execution schemes~\citep{li2026policy,xia2026dash}, the teacher supplies only a short hint
$z_{t^\star}$.  Starting from the same restored state, the same frozen student generates two
counterfactual branches, without and with the hint:
\begin{align}
    x^{\mathrm B},\xi^{\mathrm B}
    &\sim\bar\pi_\theta(\cdot\mid h_{t^\star}),\\
    x^{\mathrm H},\xi^{\mathrm H}
    &\sim\bar\pi_\theta(\cdot\mid h_{t^\star},z_{t^\star}).
    \label{eq:gi-paired-branches}
\end{align}
Because every environment action in both branches is selected by the student, the comparison remains
close to the student's rollout distribution and tests whether the current student can realize the
guidance, rather than whether the teacher can execute a better action.  With a continuation evaluator
$V$, we define
\begin{equation}
    \Delta_{t^\star}^{\mathrm{future}}
    =V(\xi^{\mathrm H})-V(\xi^{\mathrm B}),
    \qquad
    g_{t^\star}
    =\mathbf 1[\Delta_{t^\star}^{\mathrm{future}}>0].
    \label{eq:future-validation-gate}
\end{equation}
Our current instantiation uses the teacher's length-normalized likelihood of the realized continuation,
\begin{equation}
    V(\xi)
    =\frac{1}{|\mathcal A(\xi)|}
      \sum_{j\in\mathcal A(\xi)}
      \log\pi_{\mathrm T}(x_j\mid c_j),
    \label{eq:future-evaluator}
\end{equation}
where $\mathcal A(\xi)$ contains the continuation's action/content tokens.

\subsection{Hierarchical Refinement}

\paragraph{Guidance internalization.}
Validation answers whether the hint elicits a better future, but the hint is unavailable at
deployment.  We therefore transfer the induced behavior across contexts: generate with privileged
guidance, but learn to reproduce it without that guidance.  For the hinted response
$x^{\mathrm H}=(x_1^{\mathrm H},\ldots,x_M^{\mathrm H})$, define the frozen privileged target and
hint-free learner as
\begin{equation}
\begin{aligned}
    q_i^{\mathrm H}(\cdot)
    &=\operatorname{sg}\!\left[
      \bar\pi_\theta(\cdot\mid h_{t^\star},z_{t^\star},x_{<i}^{\mathrm H})
      \right],\\
    p_i^0(\cdot)
    &=\pi_\theta(\cdot\mid h_{t^\star},x_{<i}^{\mathrm H}).
\end{aligned}
\label{eq:gi-distributions}
\end{equation}
The guidance-internalization objective is
\begin{equation}
    \mathcal L_{\mathrm{GI}}
    =g_{t^\star}\frac{1}{M}
      \sum_{i=1}^{M}
      D_{\mathrm{KL}}\!\left(q_i^{\mathrm H}\,\|\,p_i^0\right).
    \label{eq:gi-objective}
\end{equation}
Because the learner conditions only on the canonical hint-free context, the accepted behavior is
transferred into the deployable policy rather than retained as a dependence on privileged guidance.

\paragraph{Confident-disagreement token focusing.}

Within a future-validated turn, we further identify token positions that provide clear corrective
supervision.  We prioritize positions where the student strongly resists the teacher candidate while
the teacher expresses a decisive preference.

For content token $i$ of the original on-policy response, let $v_i^{\mathrm S}$ and
$v_i^{\mathrm T}$ be the student and teacher top-1 candidates under the same context
$c_{t^\star,i}$.  We measure the student's counter-preference and the teacher's decisiveness as
\begin{equation}
\begin{aligned}
    g_i^{\mathrm S}
    &=\Big[
      \log\bar\pi_\theta(v_i^{\mathrm S}\mid c_{t^\star,i})
      -\log\bar\pi_\theta(v_i^{\mathrm T}\mid c_{t^\star,i})
      \Big]_+,\\
    \rho_i^{\mathrm T}
    &=\sigma\!\left(
      \log\pi_{\mathrm T}(v_i^{\mathrm T}\mid c_{t^\star,i})
      -\log\pi_{\mathrm T}(v_i^{\mathrm S}\mid c_{t^\star,i})
      \right).
\end{aligned}
\label{eq:confident-disagreement-components}
\end{equation}
Student disagreement alone can amplify ambiguous differences; teacher confidence alone can
emphasize positions the student already matches.  Their product isolates decisive corrective
conflicts,
\begin{equation}
    s_i=g_i^{\mathrm S}\rho_i^{\mathrm T},
    \qquad
    \bar r_i=\min\!\left(c_{\mathrm{tok}},1+\beta s_i\right),
    \qquad
    r_i=\frac{\bar r_i}
    {\frac{1}{|\mathcal C_{t^\star}|}
     \sum_{j\in\mathcal C_{t^\star}}\bar r_j+\epsilon}.
    \label{eq:confident-disagreement-token-weight}
\end{equation}
where $c_{\mathrm{tok}}$ caps the raw focusing weight.  For every other trained position
$i\notin\mathcal C_{t^\star}$, we set $r_i=1$.  We apply this redistribution conditionally as
$\widetilde r_{t,i}=1+\mathbf 1[t=t^\star]g_{t^\star}(r_i-1)$, preserving the supervision budget
while concentrating it on decisive corrective conflicts only within a validated turn.

\subsection{Unified Objective}

Let $m_{t,i}$ be the action/content-token mask and
$\ell_{t,i}^{\mathrm{OPD}}$ the token-level estimator used by the underlying OPD algorithm.  The
focused on-policy loss is
\begin{equation}
    \mathcal L_{\mathrm{focus}}
    =\frac{
      \sum_{t,i}m_{t,i}\widetilde r_{t,i}\ell_{t,i}^{\mathrm{OPD}}
    }{
      \sum_{t,i}m_{t,i}\widetilde r_{t,i}+\epsilon
    }.
    \label{eq:focused-opd}
\end{equation}
LENS-OPD then optimizes
$\mathcal L_{\mathrm{LENS}}=\mathcal L_{\mathrm{focus}}+
\lambda_{\mathrm{GI}}\mathcal L_{\mathrm{GI}}$.
The two terms share the future-validation gate but operate on different supervision sources:
$\mathcal L_{\mathrm{focus}}$ reweights the original on-policy response, whereas
$\mathcal L_{\mathrm{GI}}$ transfers the hint-induced response into a hint-free policy.  The complete
training procedure is provided in Algorithm~\ref{alg:lens-opd} in the Appendix~\ref{app:method_details}.
\section{Experiments}
\label{sec:experiments}

\subsection{Experimental Setup}

We evaluate LENS-OPD on ALFWorld~\citep{DBLP:conf/iclr/ShridharYCBTH21},
WebShop~\citep{DBLP:conf/nips/Yao0YN22}, and
ScienceWorld~\citep{DBLP:conf/emnlp/WangJCA22}.  Our primary setting distills a frozen
Qwen3-32B teacher into Qwen3-1.7B and Qwen3-4B students; an additional ALFWorld setting
uses a Qwen2.5-7B-RL teacher and a Qwen2.5-3B-Instruct student.  We compare against vanilla
OPD~\citep{agarwal2024policy}, TCOD~\citep{wang2026tcod},
Guided-OPD~\citep{li2026policy}, and FutureBridge-OPD (FTB)~\citep{chen2026look} under matched
training and evaluation budgets.  We report mean $\pm$ standard deviation over five random
seeds.  Dataset splits, metrics, baselines, and complete training and evaluation configurations
are provided in Appendix~\ref{app:experimental-details}, with prompt templates in
Appendix~\ref{app:prompts}.

\subsection{Main Results}

\begin{table*}[t]
  \centering
  \caption{Main results on ALFWorld, WebShop, and ScienceWorld under two Qwen3 student scales. SR denotes success rate; higher is better.}
  \label{tab:main-results}
  \setlength{\tabcolsep}{5.2pt}
  \renewcommand{\arraystretch}{1.13}
  \resizebox{\textwidth}{!}{%
  \begin{tabular}{lccccc}
    \toprule
    \textbf{Method}
      & \multicolumn{1}{c}{\textbf{ALFWorld}}
      & \multicolumn{2}{c}{\textbf{WebShop}}
      & \multicolumn{2}{c}{\textbf{ScienceWorld}} \\
    \cmidrule(lr){2-2}\cmidrule(lr){3-4}\cmidrule(lr){5-6}
      & \textbf{SR $\uparrow$}
      & \textbf{Score $\uparrow$}
      & \textbf{SR $\uparrow$}
      & \textbf{Score $\uparrow$}
      & \textbf{SR $\uparrow$} \\
    \midrule
    \rowcolor{groupGray}
    \multicolumn{6}{c}{\textbf{Qwen3-32B teacher $\rightarrow$ Qwen3-1.7B student}} \\
    Student (zero-shot)       & 5.8          & 32.7          & 5.0          & 3.7           & 0.4 \\
    Teacher (zero-shot)       & 49.3         & 51.8          & 19.0          & 44.8          & 28.7 \\
    OPD                       & $36.4\pm2.3$ & $35.8\pm13.9$ & $9.6\pm5.3$ & \cellcolor{secondSand}$42.1\pm0.3$ & $14.3\pm0.1$ \\
    TCOD-F2B                  & \cellcolor{secondSand}$42.6\pm1.4$\gain{6.2}  & $30.4\pm1.8$\loss{5.4}  & $7.4\pm1.3$\loss{2.2}  & $29.4\pm0.1$\loss{12.7} & $15.4\pm0.2$\gain{1.1} \\
    TCOD-B2F                  & $31.5\pm0.6$\loss{4.9} & $34.7\pm7.0$\loss{1.1} & $8.6\pm4.8$\loss{1.0} & \cellcolor{bestLavender}$\mathbf{42.6\pm0.5}$\gain{0.5} & \cellcolor{secondSand}$15.6\pm1.1$\gain{1.3} \\
    Guided-OPD                & $34.4\pm2.3$\loss{2.0} & $42.0\pm4.8$\gain{6.2} & $10.2\pm1.2$\gain{0.6} & $29.1\pm0.1$\loss{13.0} & $11.0\pm0.1$\loss{3.3} \\
    FTB                       & $38.2\pm4.0$\gain{1.8} & \cellcolor{secondSand}$49.5\pm1.9$\gain{13.7} & \cellcolor{bestLavender}$\mathbf{18.2\pm3.7}$\gain{8.6} & $31.7\pm0.1$\loss{10.4} & $7.7\pm0.2$\loss{6.6} \\
    \textbf{LENS-OPD (Ours)} & \cellcolor{bestLavender}$\mathbf{47.9\pm2.2}$\gain{11.5}  & \cellcolor{bestLavender}$\mathbf{49.7\pm2.2}$\gain{13.9}  & \cellcolor{secondSand}$14.4\pm5.8$\gain{4.8}  & $35.8\pm0.7$\loss{6.3}  & \cellcolor{bestLavender}$\mathbf{16.9\pm0.2}$\gain{2.6} \\
    \midrule
    \rowcolor{groupGray}
    \multicolumn{6}{c}{\textbf{Qwen3-32B teacher $\rightarrow$ Qwen3-4B student}} \\
    Student (zero-shot)       & 37.6         & 27.8          & 6.0          & 31.0          & 13.9 \\
    Teacher (zero-shot)       & 49.3         & 51.8          & 19.0          & 44.8          & 28.7 \\
    OPD                       & $42.8\pm1.3$ & $53.3\pm2.9$ & $18.8\pm0.8$ & $40.3\pm0.2$ & $20.5\pm0.2$ \\
    TCOD-F2B                  & \cellcolor{secondSand}$51.7\pm2.5$\gain{8.9}  & $50.8\pm0.6$\loss{2.5}  & \cellcolor{secondSand}$20.4\pm0.8$\gain{1.6}  & \cellcolor{bestLavender}$\mathbf{43.3\pm0.3}$\gain{3.0} & $23.3\pm0.2$\gain{2.8} \\
    TCOD-B2F                  & $43.5\pm6.0$\gain{0.7}  & $49.1\pm11.9$\loss{4.2} & $20.2\pm2.2$\gain{1.4} & $39.8\pm0.2$\loss{0.5} & $21.9\pm0.4$\gain{1.4} \\
    Guided-OPD                & $46.1\pm0.6$\gain{3.3}  & \cellcolor{secondSand}$54.3\pm3.3$\gain{1.0}  & $20.2\pm3.2$\gain{1.4}  & $38.8\pm0.1$\loss{1.5} & $14.7\pm0.4$\loss{5.8} \\
    FTB                       & $45.2\pm3.5$\gain{2.4} & $52.0\pm4.3$\loss{1.3} & $18.8\pm0.7$ & $40.9\pm0.3$\gain{0.6} & \cellcolor{bestLavender}$\mathbf{26.8\pm0.3}$\gain{6.3} \\
    \textbf{LENS-OPD (Ours)} & \cellcolor{bestLavender}$\mathbf{52.4\pm0.3}$\gain{9.6}  & \cellcolor{bestLavender}$\mathbf{54.7\pm1.8}$\gain{1.4}  & \cellcolor{bestLavender}$\mathbf{24.5\pm0.8}$\gain{5.7}  & \cellcolor{secondSand}$41.2\pm1.1$\gain{0.9}  & \cellcolor{secondSand}$23.4\pm1.1$\gain{2.9} \\
    \bottomrule
  \end{tabular}%
  }
\end{table*}

\paragraph{Overall performance.}
Table~\ref{tab:main-results} compares LENS-OPD with vanilla OPD and recent curriculum- and
selection-based baselines across three benchmarks and two student scales.  LENS-OPD delivers
strong gains across these settings.  On ALFWorld, it achieves
$47.9$ SR with the Qwen3-1.7B student, improving vanilla OPD by $11.5$ points and the strongest
baseline by $5.3$ points.  The gain remains substantial for the Qwen3-4B student, where LENS-OPD
reaches $52.4$ SR, a $9.6$-point improvement over vanilla OPD.  On WebShop, the 4B student obtains
$24.5$ SR, outperforming vanilla OPD by $5.7$ points.  On ScienceWorld, LENS-OPD improves SR from
$14.3$ to $16.9$ for the 1.7B student and from $20.5$ to $23.4$ for the 4B student.  These gains
show that hierarchical supervision allocation provides consistently effective transfer across
tasks and student scales.

\paragraph{Cross-family transfer.}

We further evaluate whether this consistency transfers to a second model family.
Table~\ref{tab:qwen25-alfworld} reports success rate and interaction efficiency on the seen, unseen,
and hard ALFWorld splits for a Qwen2.5-7B-RL teacher paired with a Qwen2.5-3B student.

\begin{table*}[t]
  \centering
  \caption{Results on ALFWorld with a Qwen2.5-7B-RL teacher and a Qwen2.5-3B student.}
  \label{tab:qwen25-alfworld}
  \setlength{\tabcolsep}{5.2pt}
  \renewcommand{\arraystretch}{1.13}
  \resizebox{\textwidth}{!}{%
  \begin{tabular}{lcccccc}
    \toprule
    \textbf{Method}
      & \multicolumn{2}{c}{\textbf{Valid Seen}}
      & \multicolumn{2}{c}{\textbf{Valid Unseen}}
      & \multicolumn{2}{c}{\textbf{Hard}} \\
    \cmidrule(lr){2-3}\cmidrule(lr){4-5}\cmidrule(lr){6-7}
      & \textbf{SR $\uparrow$}
      & \textbf{Rounds $\downarrow$}
      & \textbf{SR $\uparrow$}
      & \textbf{Rounds $\downarrow$}
      & \textbf{SR $\uparrow$}
      & \textbf{Rounds $\downarrow$} \\
    \midrule
    \rowcolor{groupGray}
    \multicolumn{7}{c}{\textbf{Qwen2.5-7B-RL teacher $\rightarrow$ Qwen2.5-3B student}} \\
    Student (zero-shot)       & 12.9 & 28.1 & 8.2 & 29.1 & 1.7 & 29.7 \\
    Teacher (zero-shot)       & 71.4 & 16.6 & 72.4 & 17.0 & 16.5 & 28.4 \\
    OPD                       & $40.4\pm3.6$ & 21.7 & \cellcolor{secondSand}$36.3\pm3.4$ & 22.9 & $6.3\pm0.9$ & 29.1 \\
    Guided-OPD                & $22.3\pm2.8$\loss{18.1} & $26.0$\roundloss{4.3} & $18.2\pm2.1$\loss{18.1} & $26.9$\roundloss{4.0} & $1.0\pm0.7$\loss{5.3} & $29.9$\roundloss{0.8} \\
    FTB                       & $25.6\pm1.9$\loss{14.8} & $25.1$\roundloss{3.4} & $22.4\pm2.2$\loss{13.9} & $26.0$\roundloss{3.1} & $1.0\pm1.4$\loss{5.3} & $29.9$\roundloss{0.8} \\
    TCOD-F2B                  & $39.5\pm3.5$\loss{0.9}
                              & $21.4$\roundgain{0.3}
                              & $32.5\pm2.7$\loss{3.8}
                              & $23.2$\roundloss{0.3}
                              & $6.6\pm1.6$\gain{0.3} & $29.1$ \\
    TCOD-B2F                  & \cellcolor{secondSand}$40.7\pm2.6$\gain{0.3}
                              & \cellcolor{secondSand}$21.0$\roundgain{0.7}
                              & $35.1\pm1.9$\loss{1.2}
                              & \cellcolor{secondSand}$22.3$\roundgain{0.6}
                              & \cellcolor{secondSand}$7.4\pm1.1$\gain{1.1} & \cellcolor{secondSand}$28.8$\roundgain{0.3} \\
    \textbf{LENS-OPD (Ours)} & \cellcolor{bestLavender}$\mathbf{46.3\pm1.8}$\gain{5.9}
                              & \cellcolor{bestLavender}$\mathbf{20.2}$\roundgain{1.5}
                              & \cellcolor{bestLavender}$\mathbf{41.9\pm2.3}$\gain{5.6}
                              & \cellcolor{bestLavender}$\mathbf{21.6}$\roundgain{1.3}
                              & \cellcolor{bestLavender}$\mathbf{9.8\pm1.4}$\gain{3.5}
                              & \cellcolor{bestLavender}$\mathbf{28.5}$\roundgain{0.6} \\
    \bottomrule
  \end{tabular}%
  }
\end{table*}

Under the Qwen2.5 teacher--student pair, LENS-OPD achieves $46.3$, $41.9$, and $9.8$ SR on Valid
Seen, Valid Unseen, and Hard, improving vanilla OPD by $5.9$, $5.6$, and $3.5$ points,
respectively.  It simultaneously reduces the average number of interaction rounds by $1.5$, $1.3$,
and $0.6$ on the three splits, and consistently improves upon both fixed TCOD curricula.  Together
with the Qwen3 results, these gains demonstrate stable transfer across benchmarks, student scales,
and model families rather than dependence on a particular backbone or predefined trajectory
ordering.

\subsection{Adaptive Supervision Dynamics}
\label{sec:adaptive-supervision-dynamics}

\begin{figure*}[t]
  \centering
  \begin{minipage}[t]{0.495\textwidth}
    \centering
    \includegraphics[width=\linewidth]{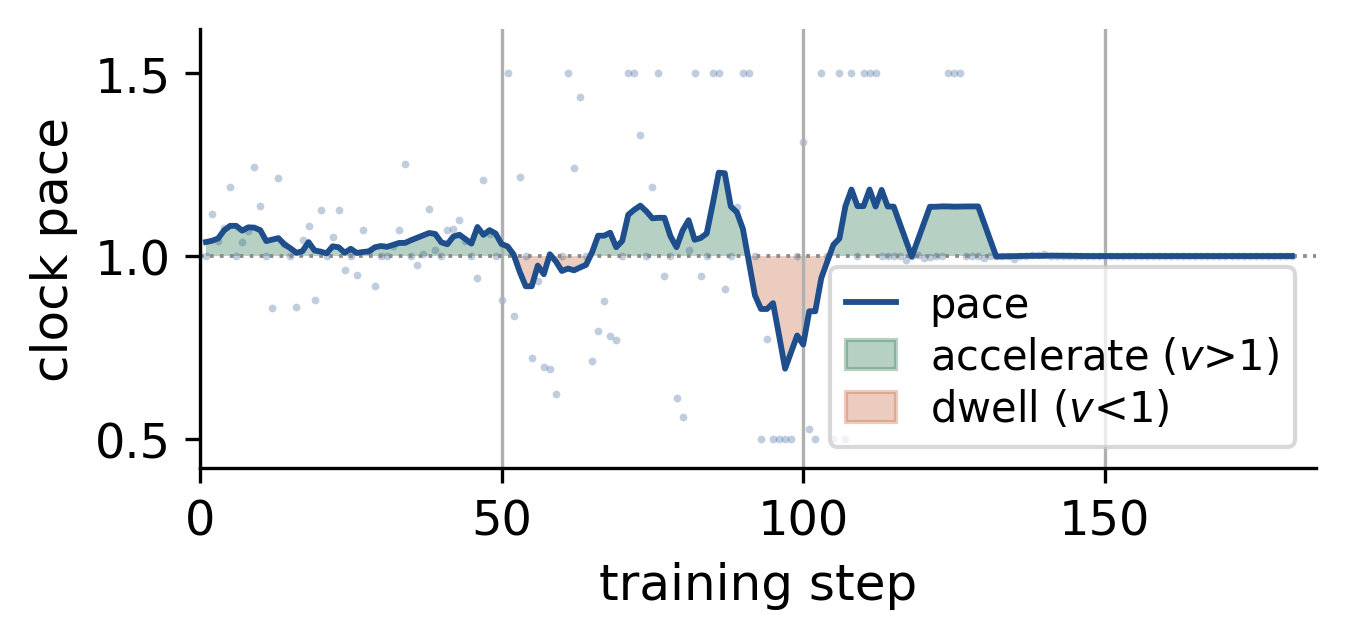}
    \vspace{-2mm}
    \centerline{\small (a) Gap-adaptive clock pace}
  \end{minipage}\hfill
  \begin{minipage}[t]{0.495\textwidth}
    \centering
    \includegraphics[width=\linewidth]{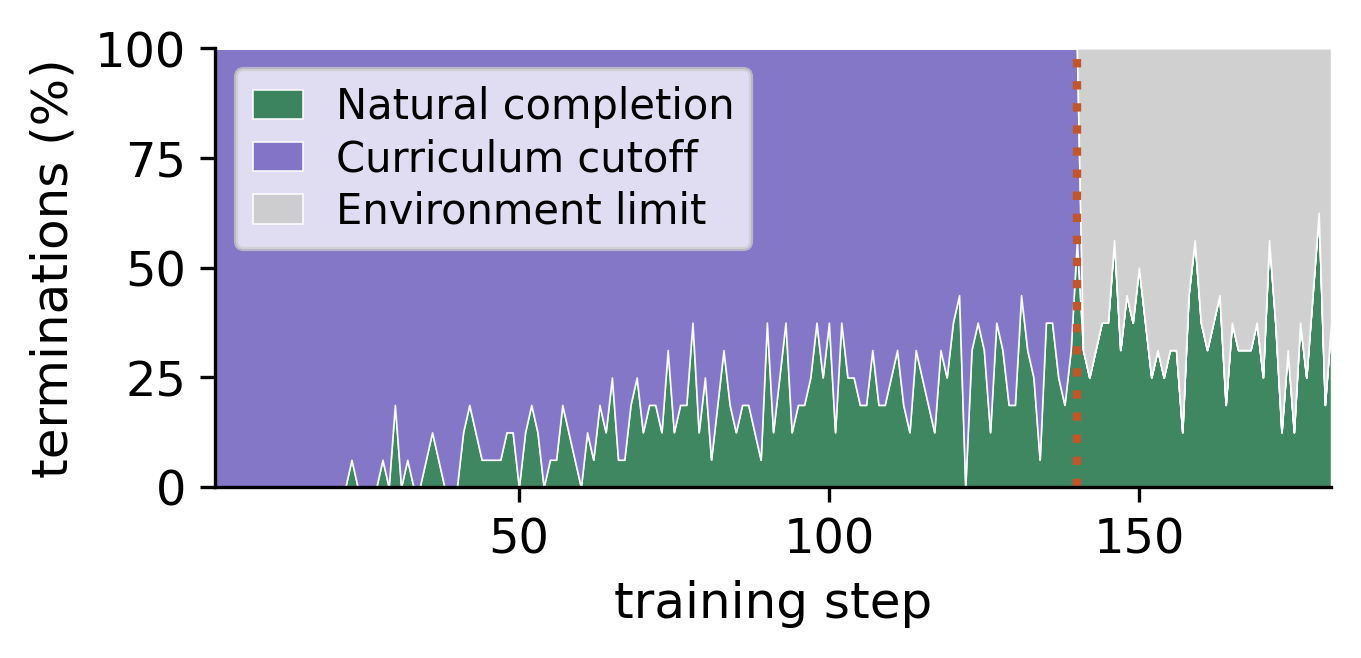}
    \vspace{-2mm}
    \centerline{\small (b) Trajectory termination modes}
  \end{minipage}
  \caption{\textbf{Gap-adaptive trajectory curriculum.} (a) Clock pace relative to fixed F2B
  ($v=1$). (b) Trajectory termination modes as the accessible horizon expands.}
  \label{fig:curriculum-dynamics}
   \vspace{-2mm}
\end{figure*}

\paragraph{The trajectory frontier adapts to the student.}
Using the Qwen3-32B teacher and Qwen3-1.7B student on ALFWorld throughout this analysis,
Figure~\ref{fig:curriculum-dynamics} shows both the controller's instantaneous pace and the resulting
trajectory termination modes.  The pace repeatedly rises above and falls below the fixed-F2B
reference $v=1$, including a pronounced slowdown near step 100, rather than following a uniform
schedule.  As the accessible horizon expands, curriculum cutoffs gradually give way to natural task
completion; after the full horizon is reached, remaining failures are governed by the environment
limit rather than curriculum truncation.

\begin{figure*}[t]
  \centering
  \begin{minipage}[t]{0.495\textwidth}
    \centering
    \includegraphics[width=\linewidth]{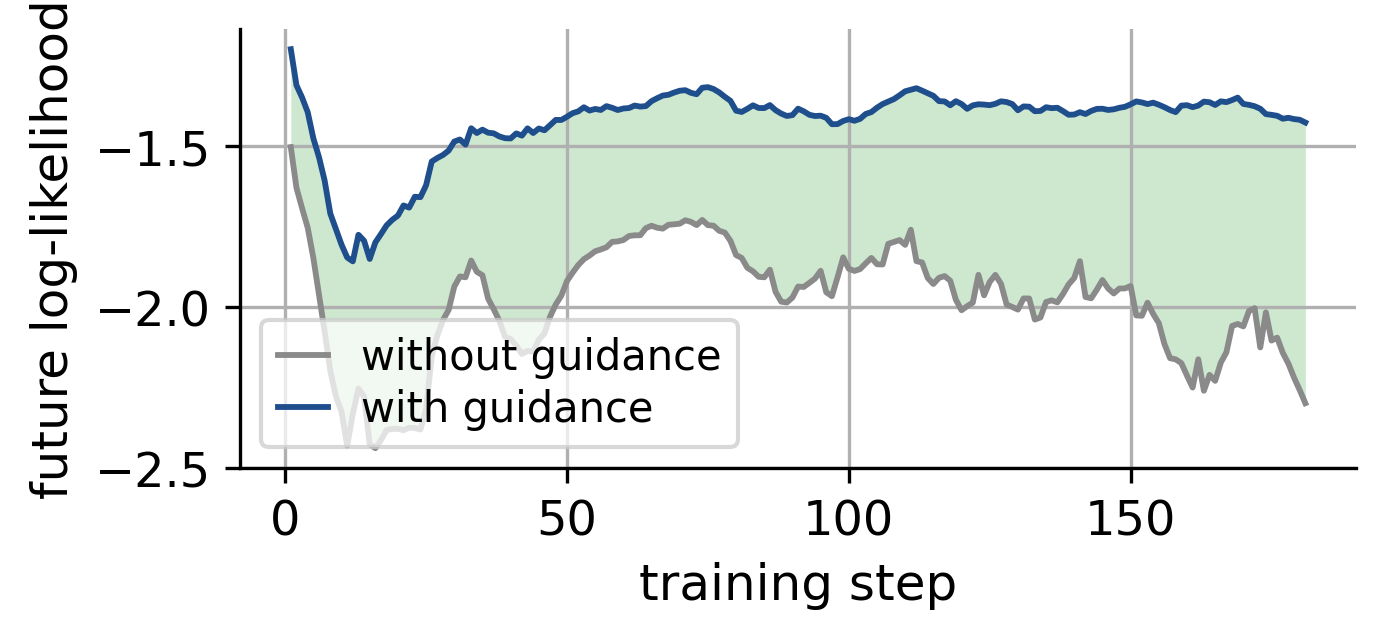}
    \vspace{-2mm}
    \centerline{\small (a) Batch-level future behavior}
  \end{minipage}\hfill
  \begin{minipage}[t]{0.495\textwidth}
    \centering
    \includegraphics[width=\linewidth]{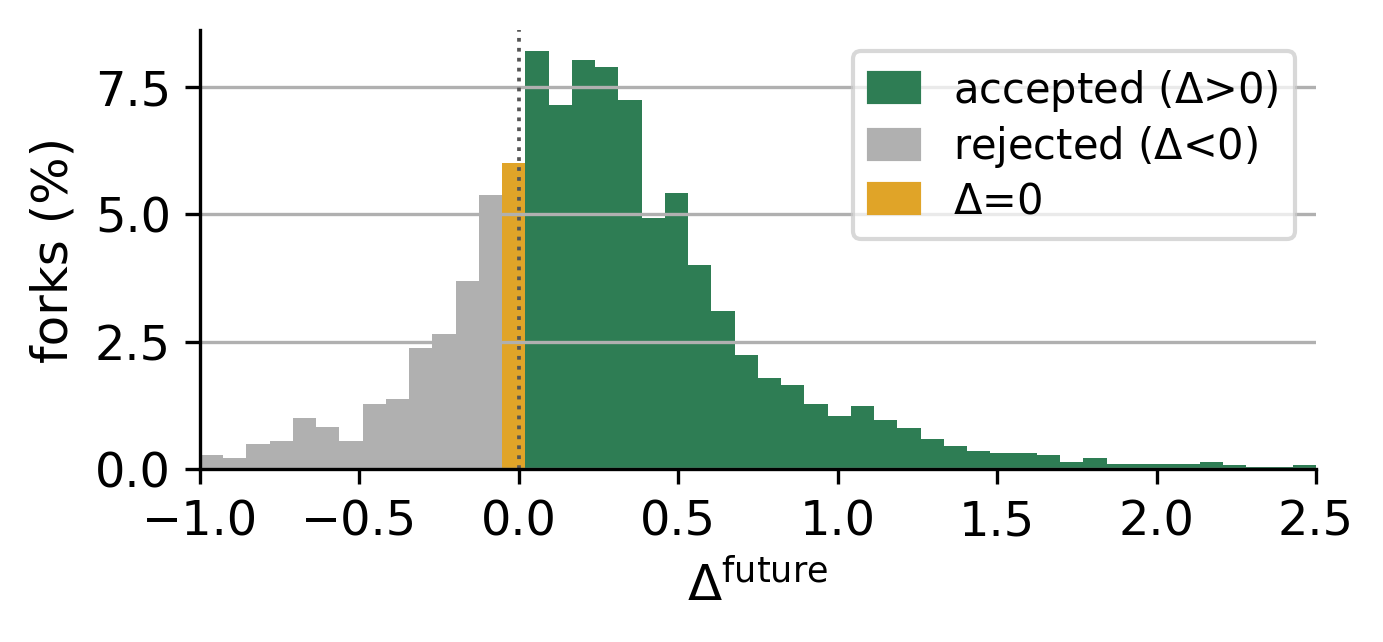}
    \vspace{-2mm}
    \centerline{\small (b) Turn-level validation outcomes}
  \end{minipage}

  \vspace{2mm}
  \begin{minipage}[t]{0.495\textwidth}
    \centering
    \includegraphics[width=\linewidth]{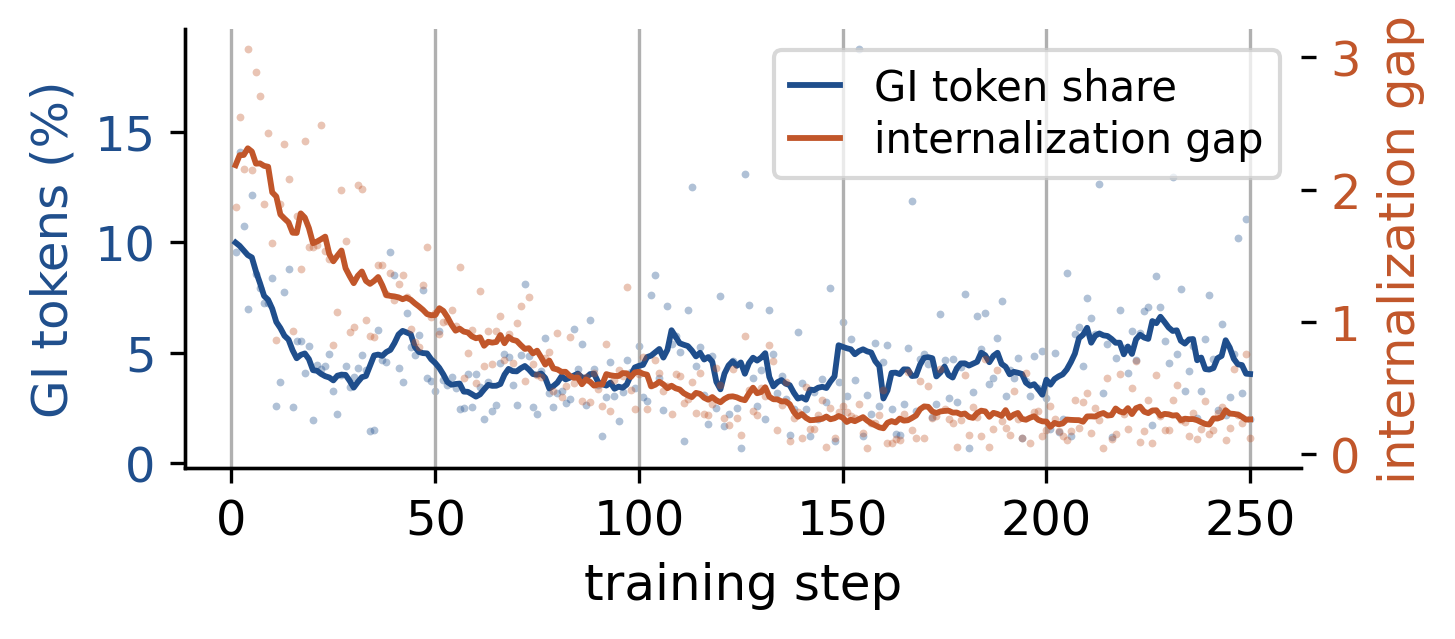}
    \vspace{-2mm}
    \centerline{\small (c) Guided-experience internalization}
  \end{minipage}\hfill
  \begin{minipage}[t]{0.495\textwidth}
    \centering
    \includegraphics[width=\linewidth]{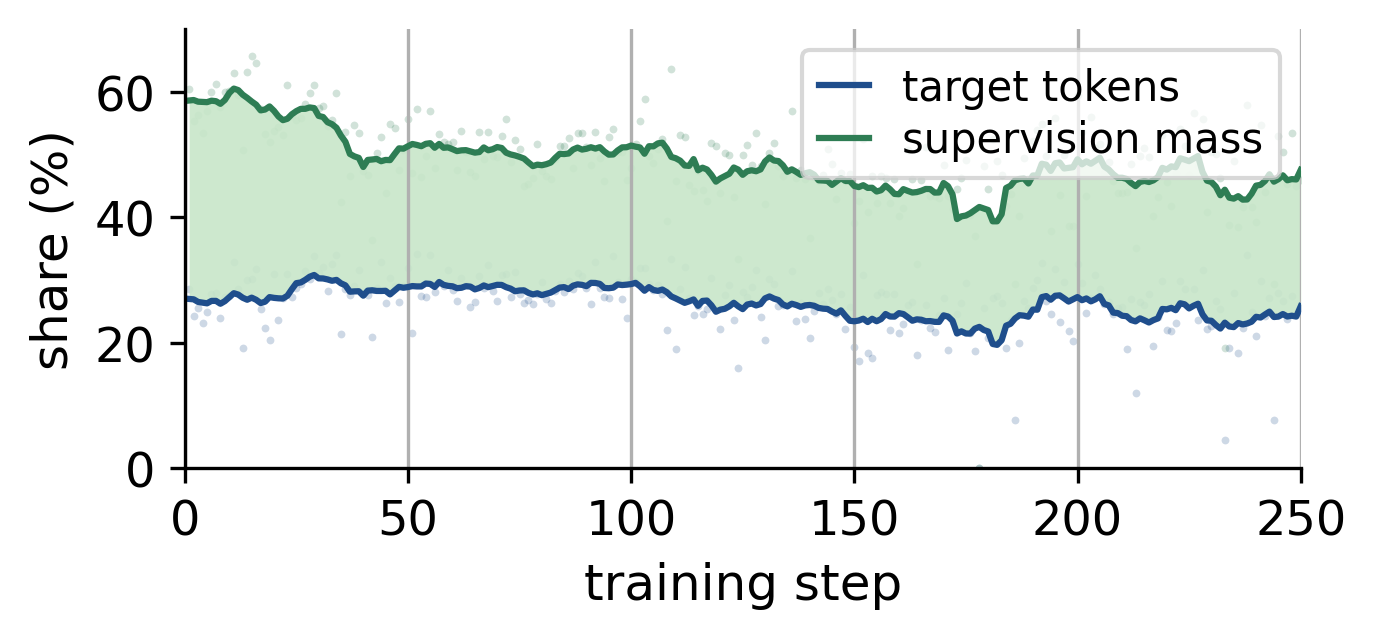}
    \vspace{-2mm}
    \centerline{\small (d) Token-level supervision concentration}
  \end{minipage}
  \caption{\textbf{Adaptive supervision dynamics.} Future validation (a,b), guided-experience
  internalization (c), and token-level supervision allocation (d).}
  \label{fig:allocation-dynamics}
  \vspace{-2mm}
\end{figure*}

\paragraph{Future validation is beneficial on average but selective across turns.}
In Figure~\ref{fig:allocation-dynamics}(a), guidance maintains a positive average future-score
advantage throughout training.  This average does not imply that every candidate is useful: the
turn-level gains in Figure~\ref{fig:allocation-dynamics}(b) span both sides of zero.  The validation
gate therefore performs a substantive selection, retaining positive-gain interventions and returning
non-positive candidates to baseline OPD.

\paragraph{Validated guidance is internalized and focused.}
Although guided experience occupies a small and gradually decreasing fraction of the training
tokens in Figure~\ref{fig:allocation-dynamics}(c), the internalization gap contracts sharply,
indicating that behaviors first elicited with guidance become increasingly reproducible by the
deployable student policy.  Within validated turns,
Figure~\ref{fig:allocation-dynamics}(d) shows a second level of allocation: target disagreement
tokens account for only roughly one quarter of the tokens but consistently receive a much larger
share of the normalized supervision mass.  Thus, LENS-OPD adapts not only how far the student
explores and which turns are reinforced, but also where learning capacity is concentrated within
those turns.

\begin{table*}[t]
  \centering
  \caption{Ablation study on ALFWorld with a Qwen3-32B teacher $\rightarrow$ Qwen3-1.7B student.}
  \label{tab:ablation}
  \setlength{\tabcolsep}{5.2pt}
  \renewcommand{\arraystretch}{1.13}
  \resizebox{\textwidth}{!}{%
  \begin{tabular}{lcccccc}
    \toprule
    \textbf{Method}
      & \multicolumn{2}{c}{\textbf{Valid Seen}}
      & \multicolumn{2}{c}{\textbf{Valid Unseen}}
      & \multicolumn{2}{c}{\textbf{Hard}} \\
    \cmidrule(lr){2-3}\cmidrule(lr){4-5}\cmidrule(lr){6-7}
      & \textbf{SR $\uparrow$}
      & \textbf{Rounds $\downarrow$}
      & \textbf{SR $\uparrow$}
      & \textbf{Rounds $\downarrow$}
      & \textbf{SR $\uparrow$}
      & \textbf{Rounds $\downarrow$} \\
    \midrule
    Fixed F2B Curriculum
      & $37.7\pm1.9$ & 23.4
      & $40.9\pm1.7$ & 23.4
      & $16.4\pm2.9$ & \cellcolor{bestLavender}$\mathbf{28.0}$ \\
    Last-Turn Proposal
      & $37.4\pm1.9$ & 24.0
      & \cellcolor{secondSand}$45.1\pm3.0$ & 23.6
      & $13.7\pm1.3$ & 28.5 \\
    w/o Guidance Internalization
      & $36.6\pm2.0$ & \cellcolor{secondSand}23.0
      & $44.8\pm2.8$ & \cellcolor{bestLavender}$\mathbf{21.9}$
      & $13.4\pm1.9$ & \cellcolor{secondSand}28.1 \\
    w/o Token Focusing
      & $39.4\pm3.7$ & 23.7
      & $43.1\pm3.4$ & 23.3
      & \cellcolor{bestLavender}$\mathbf{19.7\pm1.8}$ & 28.4 \\
    Reconsideration Prompt
      & \cellcolor{secondSand}$39.9\pm2.5$ & 23.5
      & $44.9\pm3.6$ & 22.9
      & \cellcolor{secondSand}$19.5\pm1.6$ & \cellcolor{secondSand}28.1 \\
    \midrule
    \textbf{LENS-OPD (Full)}
      & \cellcolor{bestLavender}$\mathbf{44.6\pm2.9}$
      & \cellcolor{bestLavender}$\mathbf{22.5}$
      & \cellcolor{bestLavender}$\mathbf{51.3\pm3.0}$
      & \cellcolor{secondSand}22.1
      & \cellcolor{secondSand}$19.5\pm2.2$
      & \cellcolor{secondSand}28.1 \\
    \bottomrule
  \end{tabular}%
  }
\end{table*}

\subsection{Ablation Study}

Table~\ref{tab:ablation} ablates the three levels of hierarchical allocation under the same ALFWorld
configuration.

\paragraph{Trajectory-level curriculum.}
We first replace the adaptive curriculum with a predetermined forward-to-backward schedule while
leaving turn validation and token refinement unchanged.  This reduces SR by 6.9, 10.4, and 3.1
points on Valid Seen, Valid Unseen, and Hard, respectively, and requires 0.9 and 1.3 more rounds on
the two validation splits.  These results show that a fixed training schedule cannot reliably track
the student's evolving supervision frontier.

\paragraph{Turn-level proposal.}
We replace discrepancy-based localization with the last valid turn while retaining the same paired
validation and refinement mechanisms.  This reduces SR by 7.2, 6.2, and 5.8 points on Valid Seen,
Valid Unseen, and Hard, respectively, and requires 1.5 additional rounds on both validation splits.
Thus, paired validation cannot fully compensate for a systematically poor proposal; it must begin
from a plausible decision point.

\paragraph{Post-validation refinement.}
Finally, we examine how accepted guidance is absorbed by the student.  Removing guidance
internalization reduces SR by 8.0, 6.5, and 6.1 points across the three splits.  Restoring uniform
token weights lowers SR by 5.2 points on Valid Seen and 8.2 points on Valid Unseen.  Replacing the
default feedback-incorporation instruction with the \emph{Reconsideration Prompt} in
Appendix~\ref{app:student-feedback-prompts} reduces SR by 4.7 and 6.4 points on the two validation
splits, while Hard performance remains comparable.  These results show that useful guidance must be
internalized into the hint-free policy and focused on decisive token-level conflicts.

\section{Conclusion}
\label{sec:conclusion}

We formulate long-horizon OPD as hierarchical supervision allocation: productive guidance must
improve future behavior while remaining learnable by the current student.  LENS-OPD implements this
principle through \textsc{Locate}, \textsc{Validate}, and \textsc{Refine}, which adapt trajectory
exposure, verify guidance with paired student-controlled continuations, and focus learning on
validated behaviors and decisive token-level conflicts.  Across agent benchmarks, student scales,
and model families, LENS-OPD consistently improves over vanilla OPD and strong adaptive baselines,
showing the importance of jointly coordinating exploration, intervention, and internalization.

\bibliography{iclr2027_conference,related_work_dblp}
\bibliographystyle{iclr2027_conference}

\clearpage
\appendix

\section{Additional Method Details}
\label{app:method_details}
\subsection{Training Algorithm Overview}

Algorithm~\ref{alg:lens-opd} summarizes one LENS-OPD training iteration.  The three stages operate
at progressively finer resolutions: the curriculum controls trajectory exposure, future validation
evaluates one proposed intervention, and refinement converts an accepted intervention into deployable
behavioral and token-level supervision.  The curriculum statistic is measured at the current rollout
frontier, whereas the proposed intervention may occur at any valid turn within the exposed trajectory.

\begin{algorithm}[H]
\caption{One training iteration of LENS-OPD}
\label{alg:lens-opd}
\small
\begin{algorithmic}[1]
\Require Teacher $\pi_{\mathrm T}$; trainable student $\pi_\theta$; frozen rollout student
$\bar\pi_\theta$; controller state $(K,e,G_K^{\mathrm{entry}})$; residence budget $\eta$
\Ensure Updated student $\pi_\theta$ and curriculum state
\Statex \textbf{Locate: adaptive exposure and turn proposal}
\State Roll out $\bar\pi_\theta$ for at most $K$ turns to obtain a batch $\mathcal B$
\If{at least one valid episode reaches depth $K$}
    \State Compute frontier competence $C_{n,K}$ over valid episodes
    \State Initialize $G_K^{\mathrm{entry}}$ if the controller has just entered depth $K$
    \State $v\gets\operatorname{clip}\!\left(
      (G_K^{\mathrm{entry}}+\epsilon)/(1-C_{n,K}+\epsilon),0.5,1.5\right)$
\Else
    \State $v\gets 1$ \Comment{fall back to the reference F2B pace}
\EndIf
\State $e\gets e+v$; if $e\ge\eta$, set $K\gets\min(K+1,K_{\max})$ and $e\gets e-\eta$
\ForAll{student trajectories $\tau_b\in\mathcal B$}
    \State Compute mean teacher--student divergence $D_{b,t}$ for every valid turn $t$
    \State $t_b^\star\gets\arg\max_t D_{b,t}$
    \Statex \textbf{Validate: paired future continuations}
    \State Restore the canonical history and environment state at $t_b^\star$
    \State Generate base response and continuation $(x_b^{\mathrm B},\xi_b^{\mathrm B})$
    \State Query one teacher hint $z_b$ and generate $(x_b^{\mathrm H},\xi_b^{\mathrm H})$
    \State $g_b\gets\mathbb I\!\left[V(\xi_b^{\mathrm H})-V(\xi_b^{\mathrm B})>0\right]$
    \Statex \textbf{Refine: guidance internalization and token focusing}
    \State Initialize $\mathcal L_{\mathrm{GI}}^{(b)}\gets0$ and $\widetilde r_{b,t,i}\gets1$
    \If{$g_b=1$}
        \State Construct the hinted frozen-student target and hint-free learner
        \State Compute cross-context guidance-internalization loss $\mathcal L_{\mathrm{GI}}^{(b)}$
        \State Compute normalized confident-disagreement weights $r_{b,i}$ on original turn $t_b^\star$
        \State Set $\widetilde r_{b,t_b^\star,i}\gets r_{b,i}$
    \EndIf
    \State Compute $\mathcal L_{\mathrm{focus}}^{(b)}$ on the original on-policy trajectory
\EndFor
\State $\mathcal L\gets
\operatorname{Aggregate}_b\!\left[
\mathcal L_{\mathrm{focus}}^{(b)}+
\lambda_{\mathrm{GI}}\mathcal L_{\mathrm{GI}}^{(b)}\right]$
\State Update $\theta$ using $\nabla_\theta\mathcal L$ and synchronize $\bar\pi_\theta$ according to the rollout schedule
\end{algorithmic}
\end{algorithm}

\section{Experimental Details}
\label{app:experimental-details}

\subsection{Training Configuration}
We adopt the shared training configuration in TCOD~\citep{wang2026tcod}, Appendix~D.3,
with five random seeds (42, 43, 44, 45, and 46). Table~\ref{tab:training-config}
records the optimization, rollout, and distributed execution settings. Method-specific
objectives and guidance mechanisms are described in the main text. Each run uses 250
training steps, with evaluation every five steps and checkpoint saving at step 250.
For the ALFWorld experiments with the Qwen2.5-7B-RL teacher, we use
\nolinkurl{siaosiao/grpo-qwen2.5-7b-alfworld-global-step-150},
a Qwen2.5-7B RL checkpoint obtained after 150 steps of GRPO training on ALFWorld.
This checkpoint is kept frozen throughout student distillation. For LENS-OPD, the curriculum
residence budget is set to $\eta=5$, matching the fixed F2B clock when $v=1$. We validate one
Top-KL turn per trajectory using three-turn paired continuations.

Within each comparison, methods share the teacher checkpoint, student initialization, prompts,
environment configuration, action parser, and rollout, optimization, and evaluation budgets in a
common codebase. The teacher remains frozen, and student trajectories are collected on-policy at
temperature $1.0$ without beam search or candidate reranking. We compare with vanilla
OPD~\citep{agarwal2024policy}, the forward-to-backward and backward-to-forward variants of
TCOD~\citep{wang2026tcod}, Guided-OPD~\citep{li2026policy}, and
FutureBridge-OPD~\citep{chen2026look}. The zero-shot student and teacher are reported as references
and excluded from method rankings.

\begin{table}[H]
\centering
\small
\caption{Training configuration across the three benchmarks. Batch-size field names follow TCOD.}
\label{tab:training-config}
\begin{tabular}{ll}
\toprule
Parameter & Value \\
\midrule
KL coefficient & 1.0 \\
Learning rate & $1\times10^{-6}$ \\
Gradient clipping & 1.0 \\
Repeat times & 1 \\
Maximum sample staleness & 2 \\
Training steps & 250 \\
Batch size / train batch size & 16 / 64 \\
Save / evaluation interval & 250 / 5 steps \\
Maximum prompt / response tokens & 10,240 / 512 \\
Rollout temperature & 1.0 \\
Log-probability collection & All tokens \\
Random seeds & 42, 43, 44, 45, 46 \\
\midrule
Curriculum residence budget $\eta$ & 5 \\
Top-KL turn proposals per trajectory & 1 \\
Future-validation horizon $H$ & 3 turns \\
Guidance-score temperature $\tau$ & 1.0 \\
Token-focus scale $\beta$ & 1.0 \\
Raw token-weight cap $c_{\mathrm{tok}}$ & 5.0 \\
\midrule
Nodes / GPUs per node & 1 / 8 \\
GPU model & NVIDIA H100 \\
Tensor / sequence parallelism & 2 / 2 (Ulysses) \\
Maximum tokens per GPU & 16,384 \\
GPU memory utilization & 0.7 \\
Precision & BFloat16 \\
\bottomrule
\end{tabular}
\end{table}

\subsection{Evaluation Configuration}
Following TCOD and FutureBridge-OPD~\citep{wang2026tcod,chen2026look}, we use the
standard ALFWorld training and unseen-evaluation splits, and the TCOD-aligned WebShop
training subset with 100 held-out sessions.  Evaluation uses the decoding and execution
settings in TCOD Appendix~D.4, summarized in Table~\ref{tab:evaluation-config}. Thinking
mode is disabled. The evaluation generation limit is 4,096 tokens, whereas training rollouts
use a maximum response length of 512 tokens. The environment interaction limits apply during
both training and evaluation. We report the mean and standard deviation across the five seeds above.

\begin{table}[H]
\centering
\small
\caption{Evaluation decoding and environment execution settings.}
\label{tab:evaluation-config}
\begin{tabular}{ll}
\toprule
Parameter & Value \\
\midrule
Maximum generation tokens & 4,096 \\
Temperature & 0.4 \\
Top-$p$ / top-$k$ / min-$p$ & 1.0 / $-1$ / 0.0 \\
History length & 2 steps \\
Maximum interactions: ALFWorld & 30 \\
Maximum interactions: ScienceWorld & 30 \\
Maximum interactions: WebShop & 15 \\
Evaluation workers & 8 \\
Process timeout & 3,600 seconds \\
Synchronization & Dynamic by explorer \\
\bottomrule
\end{tabular}
\end{table}

\paragraph{Evaluation metrics.}
For each benchmark, let $N$ denote the number of evaluation episodes for a given seed.
ALFWorld uses the environment's task-completion criterion: with $c_i\in\{0,1\}$ indicating
whether episode $i$ successfully completes the task, we report
\begin{equation}
\mathrm{SR}_{\mathrm{ALFWorld}}=\frac{100}{N}\sum_{i=1}^{N}c_i.
\end{equation}
For WebShop, let $r_i\in[0,1]$ be the final environment reward. The score follows the
reward average used by FutureBridge-OPD~\citep{chen2026look}, while success requires
$r_i=1$:
\begin{equation}
\mathrm{Score}_{\mathrm{WebShop}}=\frac{100}{N}\sum_{i=1}^{N}r_i,
\qquad
\mathrm{SR}_{\mathrm{WebShop}}=\frac{100}{N}\sum_{i=1}^{N}\mathbf{1}[r_i=1].
\end{equation}
For ScienceWorld, let $s_i$ denote the final environment score on its native 0--100 scale,
with any negative terminal score replaced by zero before averaging. Success follows the
environment's standard task-completion criterion, denoted by $c_i\in\{0,1\}$:
\begin{equation}
\mathrm{Score}_{\mathrm{ScienceWorld}}=\frac{1}{N}\sum_{i=1}^{N}\max(s_i,0),
\qquad
\mathrm{SR}_{\mathrm{ScienceWorld}}=\frac{100}{N}\sum_{i=1}^{N}c_i.
\end{equation}
Scores use the final outcome of each episode, including episodes that reach the interaction
limit. We compute each metric separately for each seed and then report its mean and
standard deviation across the five seeds.

\section{Prompt Templates}
\label{app:prompts}

We use two distinct families of prompts.  The environment interaction prompts define the ordinary
student and teacher policies used by all OPD baselines.  They follow the benchmark-specific
templates and output conventions of TCOD Appendix~E~\citep{wang2026tcod}.
LENS-OPD additionally uses a short diagnostic cue and a student-side feedback injection only inside
the paired future-validation branch.  The cue is privileged audit information: the teacher never
executes an environment action, and neither audit continuation is inserted into the ordinary OPD
buffer.

\subsection{Environment Interaction Prompts}

The three environments share the same slot structure---task, recent interaction history, current
observation, and currently available actions---while retaining their native action vocabularies.
The templates below are functionally identical to those used by TCOD; braces denote runtime fields.

\promptheading{ALFWorld}
\begin{promptbox}
You are an expert agent operating in the ALFRED Embodied Environment. Your task is to: {task_description}
Prior to this step, you have already taken {step_count} step(s). Below are the most recent {history_length} observations and the corresponding actions you took: {action_history}
You are now at step {current_step} and your current observation is: {current_observation}
Your admissible actions of the current situation are: [{admissible_actions}].

Now it's your turn to take an action.
You should first reason step-by-step about the current situation.
Once you've finished your reasoning, you should choose an admissible action for current step and present it within <action> </action> tags.
\end{promptbox}

\promptheading{ScienceWorld}
\begin{promptbox}
Your ScienceWorld task is: {task_description}
Prior to this step, you have already taken {step_count} step(s). Below are the most recent {history_length} observations and the corresponding actions you took: {action_history}
You are now at step {current_step} and your current observation is: {current_observation}
Available action commands: [{action_templates}]
Available objects you can interact with: [{objects}]

Now it's your turn to take an action. Combine an action command with appropriate object(s) to form a valid action.
You should first reason step-by-step about the current situation.
Once you've finished your reasoning, you should choose a valid action for the current step and present it within <action> </action> tags.
\end{promptbox}

\promptheading{WebShop}
\begin{promptbox}
You are an expert autonomous agent operating in the WebShop e-commerce environment.
Your task is to: {task_description}.
Prior to this step, you have already taken {step_count} step(s). Below are the most recent {history_length} observations and the corresponding actions you took: {action_history}
You are now at step {current_step} and your current observation is: {current_observation}.
Your admissible actions of the current situation are: 
[
{available_actions}
].

Now it's your turn to take one action for the current step.
You should first reason step-by-step about the current situation, then think carefully which admissible action best advances the shopping goal.
Once you've finished your reasoning, you should choose an admissible action for current step and present it within <action> </action> tags.
\end{promptbox}
WebShop actions use the environment-native \texttt{search[query]} and
\texttt{click[element]} formats whenever those controls are available.

\subsection{Teacher Diagnostic Cue}

At the proposed fork turn, the teacher observes the same replay-restored environment state that the
student is about to face.  We provide the full admissible-action set, the student's proposed action,
and the two most recent interaction turns.  The teacher generates exactly one diagnostic sentence.

\begin{promptbox}
You are diagnosing a student agent's proposed action. You see exactly
what the student sees.

Task:
{task}

Current observation:
{observation}

Available actions:
{admissible_actions}

Interaction history:
{history}

The student is about to take this action:
{student_action}

Judge this action against the current state and give exactly one short
hint.

Rules:
- If the action is wrong, or has an important oversight/constraint the
  student should reconsider, POINT IT OUT DIRECTLY -- say what the
  issue is and what the student should do instead (you may name the
  correct action).
- If the action is right, ENCOURAGE / confirm the student: say the
  action is on the right track and (if useful) suggest the next step.
- Do not write a tool call.
- One sentence only.
\end{promptbox}

We decode the cue greedily (temperature $0$) with at most 80 generated tokens and retain the two
most recent interaction turns as history.  If cue generation fails, the trajectory receives no
forked audit and is not retried.  This asymmetric instruction deliberately supports both correction
and confirmation: a wrong proposal receives a direct diagnosis, whereas a correct proposal receives
confirmation and, when useful, a next-step suggestion.

\subsection{Student Feedback Injection and Prompt Ablation}
\label{app:student-feedback-prompts}

For the hinted branch, our default \emph{Feedback-Incorporation Prompt} is inserted at the fork turn's student-message
position.  The prefix preceding the insertion is otherwise identical to the base branch.  The cue
therefore changes only the information available to the student; the teacher still does not act.

\begin{promptbox}
You proposed the following action earlier:

{student_action}

Teacher feedback:

{teacher_hint}

Take the teacher's feedback into account and produce the next action
yourself.
\end{promptbox}

In the \emph{Reconsideration Prompt} ablation, we keep all fields and formatting fixed but replace
the final instruction with the following more forceful alternative:

\begin{promptbox}
Reconsider your decision and produce the next action yourself.
\end{promptbox}

Both variants require the student to produce the next action itself.  The ablation therefore tests
the effect of feedback framing rather than teacher execution, additional state information, or a
different action interface.

\section{Additional Results}
\label{app:additional-results}

\subsection{WebShop Training Dynamics}
\label{app:webshop-training-dynamics}

We extend the trajectory and future-validation diagnostics in
Figures~\ref{fig:curriculum-dynamics} and~\ref{fig:allocation-dynamics}(a,b) to WebShop,
using a Qwen3-32B teacher with Qwen3-1.7B and Qwen3-4B students.
Each row shows one training run. Training-step coordinates identify the student version
that generated each rollout. Clock-pace curves use centered seven-batch moving averages;
future-score curves pool completed paired validations over the same window size.
Termination proportions and gain histograms are unsmoothed.

\begin{figure}[H]
  \centering
  \includegraphics[width=\linewidth]{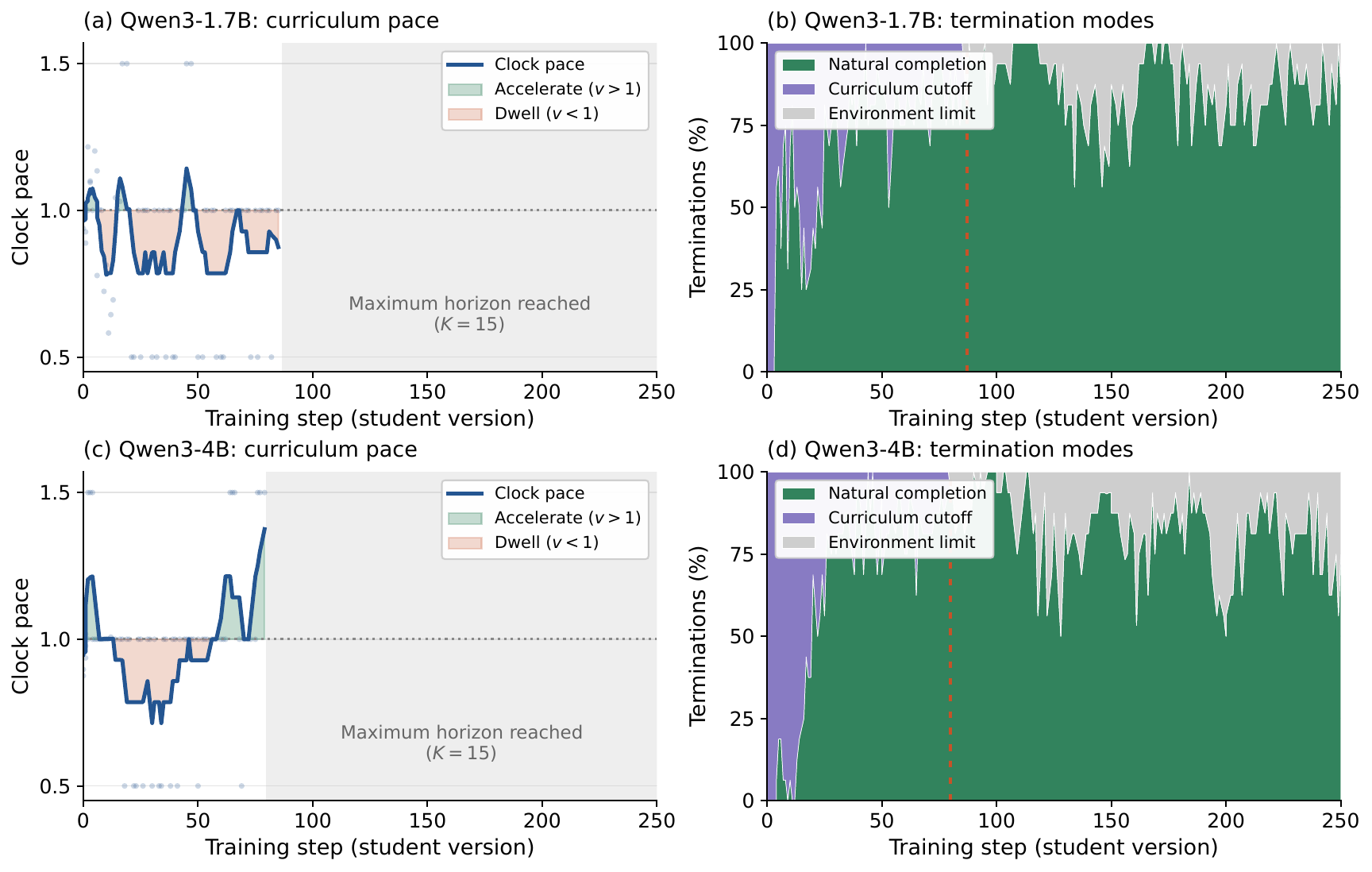}
  \caption{\textbf{WebShop trajectory curriculum.} Top: Qwen3-1.7B; bottom: Qwen3-4B.
  Left: recorded clock pace (dots) and its moving average, relative to fixed F2B
  ($v=1$). Right: trajectory-termination proportions; dashed lines mark the first
  rollouts using the full horizon.}
  \label{fig:webshop-curriculum-dynamics}
\end{figure}

Figure~\ref{fig:webshop-curriculum-dynamics} shows that both curricula adjust their pace
before reaching the full 15-turn horizon, first used at student versions 87 and 80 for
the 1.7B and 4B students, respectively. Subsequent terminations arise from natural
environment completion or the interaction limit. Natural completion does not imply
task success. The shaded region in the pace panels marks the completed curriculum;
no clock speed is imputed there.

\begin{figure}[H]
  \centering
  \includegraphics[width=\linewidth]{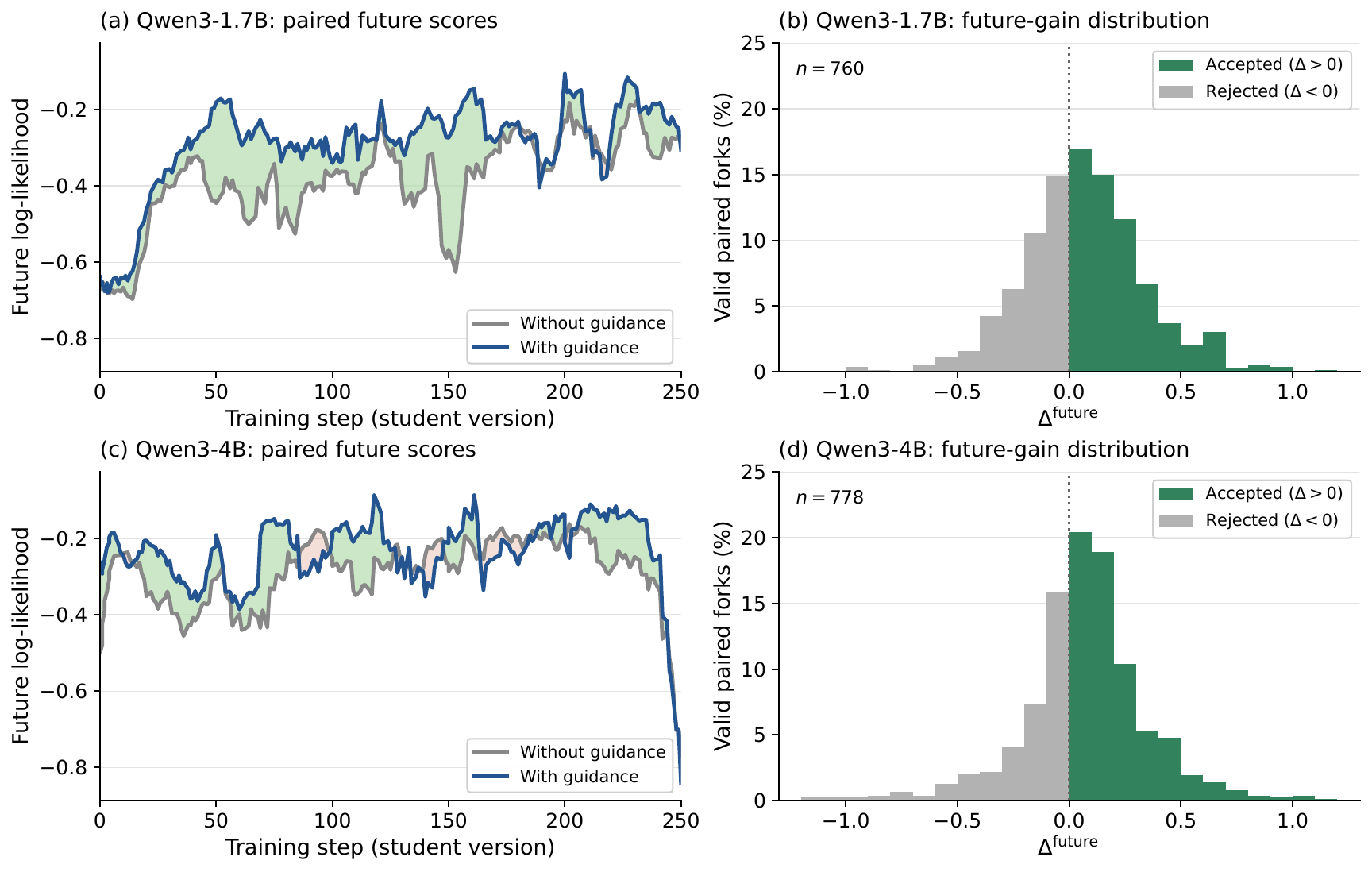}
  \caption{\textbf{WebShop future validation.} Top: Qwen3-1.7B; bottom: Qwen3-4B.
  Left: guided and unguided future scores over the same complete pairs.
  Right: unsmoothed future-gain distributions, normalized by the number of complete
  pairs in each run. Green and gray denote positive and negative gains, respectively.}
  \label{fig:webshop-future-validation}
\end{figure}

Figure~\ref{fig:webshop-future-validation} compares length-normalized teacher
log-likelihoods of guided and unguided continuations from the same fork.
There are 760 complete pairs among 3,165 recorded episodes for the 1.7B student and
778 among 3,245 for the 4B student. Incomplete pairs, predominantly due to timeouts,
are excluded rather than assigned zero gain; these comparisons therefore describe
completed validations. One unfinished trailing batch per run, containing eight
episodes without a logged student-version mapping, is excluded from both figures.
Across complete pairs, mean future gains are 0.066 and 0.069, respectively, while
both distributions retain a negative-gain region. Guidance is beneficial on average
within this observed subset, but individual interventions still require validation.

\subsection{ScienceWorld Training Dynamics}
\label{app:scienceworld-training-dynamics}

\begin{figure}[H]
  \centering
  \includegraphics[width=\linewidth]{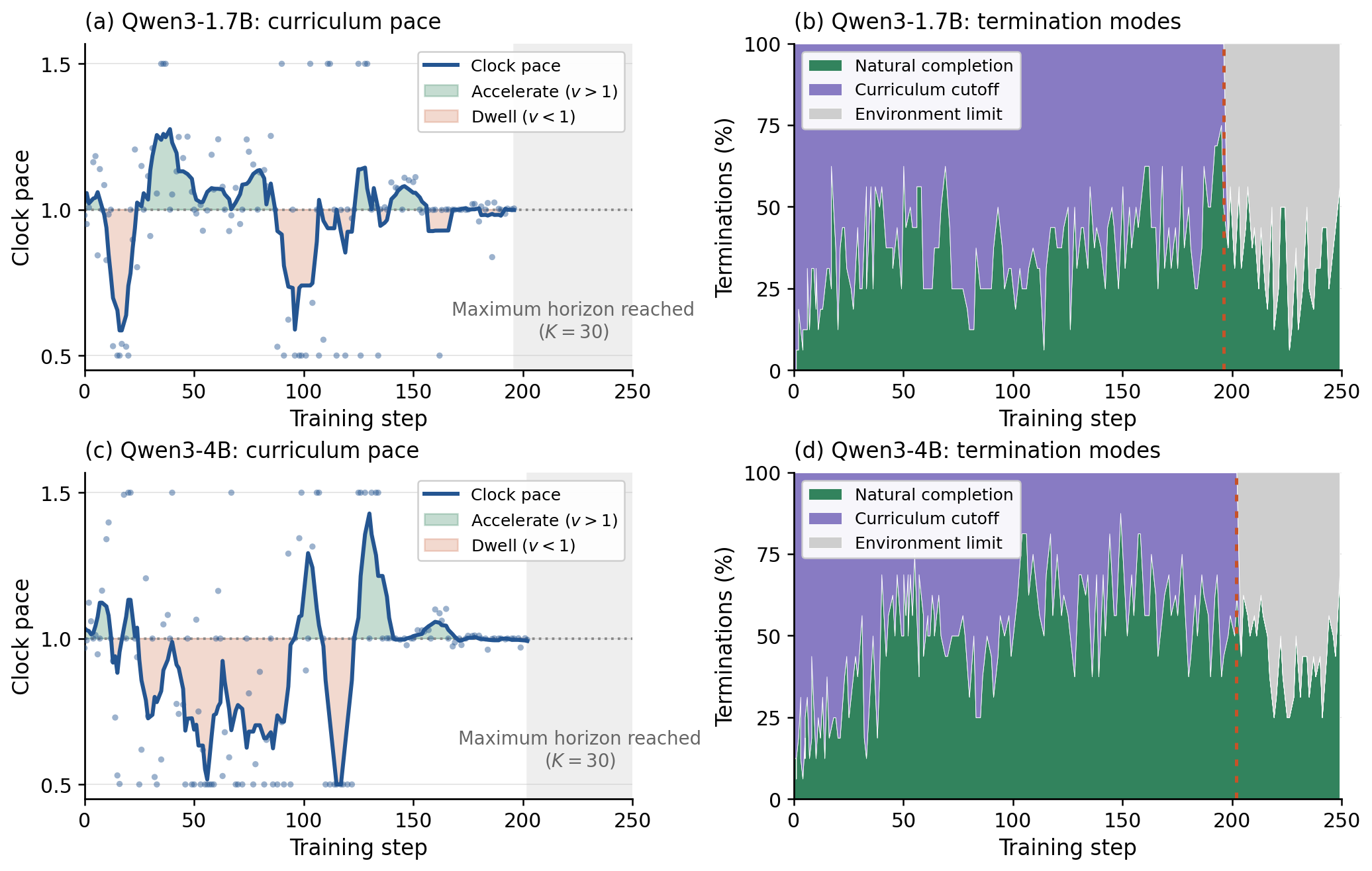}
  \caption{\textbf{ScienceWorld trajectory curriculum.} Top: Qwen3-1.7B; bottom: Qwen3-4B.
  Left: clock pace relative to fixed F2B ($v=1$). Right: trajectory-termination proportions.
  Dashed lines mark the first rollouts using the full $K=30$ horizon.}
  \label{fig:scienceworld-curriculum-dynamics}
\end{figure}

We repeat the same trajectory and future-validation diagnostics on ScienceWorld with a Qwen3-32B
teacher and Qwen3-1.7B or Qwen3-4B students.  As shown in
Figure~\ref{fig:scienceworld-curriculum-dynamics}, the clock repeatedly accelerates and dwells relative
to fixed F2B before both runs reach the full $K=30$ horizon near step 200.  Over the same period,
curriculum cutoffs recede as natural completion becomes more frequent; after the full horizon is
reached, trajectories terminate through natural completion or the environment limit.  The distinct
pace profiles at the two student scales further show that trajectory exposure follows the observed
student--teacher gap rather than a shared preset schedule.

\begin{figure}[H]
  \centering
  \includegraphics[width=\linewidth]{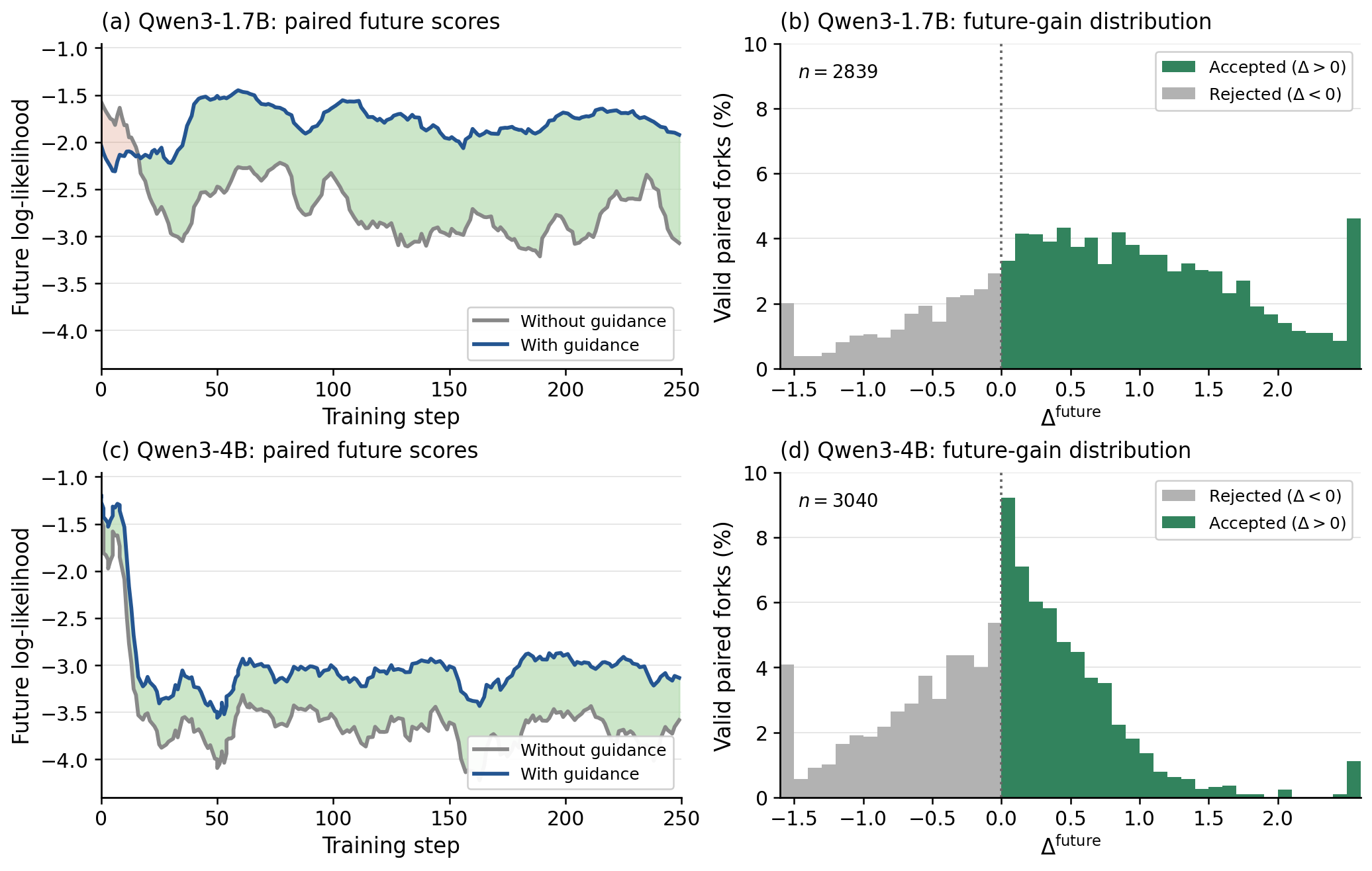}
  \caption{\textbf{ScienceWorld future validation.} Top: Qwen3-1.7B; bottom: Qwen3-4B.
  Left: guided and unguided future scores. Right: future-gain distributions over valid paired forks,
  normalized separately for each run. Green and gray denote positive and negative gains.}
  \label{fig:scienceworld-future-validation}
\end{figure}

Figure~\ref{fig:scienceworld-future-validation} shows that guided continuations achieve higher
teacher-evaluated future scores over most of training for both student scales.  Nevertheless, the
2,839 and 3,040 valid paired forks for the 1.7B and 4B students, respectively, contain gains on both
sides of zero.  Thus, the average benefit of guidance transfers to ScienceWorld, while the
turn-specific variation continues to justify paired validation rather than accepting every proposed
intervention.

\clearpage

\clearpage

\FloatBarrier

\end{document}